\documentclass[10pt,twocolumn,letterpaper]{article}

\usepackage[pagenumbers]{cvpr} % To force page numbers, e.g. for an arXiv version

\usepackage{graphicx}
\usepackage{amsmath}
\usepackage{amssymb}
\usepackage{amsfonts}
\usepackage{booktabs}
\usepackage{multirow}
\usepackage{siunitx}
\usepackage[flushleft]{threeparttable}
\usepackage{array}
\newcolumntype{C}[1]{>{\centering\arraybackslash}p{#1}}
\usepackage[all,cmtip]{xy}
\usepackage[dvipsnames, table]{xcolor}
\usepackage{color, colortbl}
\usepackage{caption}
\usepackage{subcaption}
\usepackage{adjustbox}
\usepackage[normalem]{ulem}
\useunder{\uline}{\ul}{}

\usepackage{paralist}
\usepackage{algorithm}
\usepackage{listings}

\makeatletter
\AfterEndEnvironment{algorithm}{\let\@algcomment\relax}
\AtEndEnvironment{algorithm}{\kern2pt\hrule\relax\vskip3pt\@algcomment}
\let\@algcomment\relax
\newcommand\algcomment[1]{\def\@algcomment{\footnotesize#1}}
\renewcommand\fs@ruled{\def\@fs@cfont{\bfseries}\let\@fs@capt\floatc@ruled
	\def\@fs@pre{\hrule height.8pt depth0pt \kern2pt}%
	\def\@fs@post{}%
	\def\@fs@mid{\kern2pt\hrule\kern2pt}%
	\let\@fs@iftopcapt\iftrue}
\makeatother

\usepackage[LGR,T1]{fontenc}
\DeclareMathAlphabet{\mathgtt}{LGR}{cmtt}{m}{n}

\definecolor{table-bg}{rgb}{0.96, 0.96, 0.97}

\usepackage{tabulary,multirow,overpic}
\newcommand{\tablestyle}[2]{\setlength{\tabcolsep}{#1}\renewcommand{\arraystretch}{#2}\centering}
\newlength\savewidth\newcommand\shline{\noalign{\global\savewidth\arrayrulewidth
  \global\arrayrulewidth 1.2pt}\hline\noalign{\global\arrayrulewidth\savewidth}}
\newcommand{\cgap}[2]{
\fontsize{6pt}{1em}\selectfont{(${#1}${#2})}
}
\newcommand{\smallscript}[1]{
    \color[HTML]{9B9B9B}
    \fontsize{6pt}{1em}\selectfont{{#1}}
}

\makeatletter
\renewcommand{\paragraph}{%
  \@startsection{paragraph}{4}%
  {\z@}{0.2ex \@plus 1ex \@minus .2ex}{-1em}%
  {\normalfont\normalsize\bfseries}%
}
\makeatother

\usepackage{pmboxdraw}

\usepackage[pagebackref,breaklinks,colorlinks]{hyperref}
\usepackage[capitalize]{cleveref}
\crefname{section}{Sec.}{Secs.}
\Crefname{section}{Section}{Sections}
\Crefname{table}{Table}{Tables}
\crefname{table}{Tab.}{Tabs.}

\def\cvprPaperID{1751} % *** Enter the CVPR Paper ID here
\def\confName{CVPR}
\def\confYear{2022}

\begin{document}

\setlength{\abovedisplayskip}{0.5em}
\setlength{\belowdisplayskip}{0.5em}

%%%%%%%%% TITLE - PLEASE UPDATE
% 不要使用region classification作为标题！因为与主要method不符
% 不存在pixel-region对应关系（因为region是一些pixel的集合），应当是pixel和token/image primitive的对应关系
\title{Semantic Segmentation by Early Region Proxy}

\author{Yifan Zhang \quad Bo Pang \quad Cewu Lu\thanks{Cewu Lu is corresponding author, member of Qing Yuan Research Institute and MoE Key Lab of Artificial Intelligence, AI Institute, Shanghai Jiao Tong University, China and Shanghai Qi Zhi Institute.}\\
Shanghai Jiao Tong University\\
{\tt\small \{zhangyf\_sjtu, pangbo, lucewu\}@sjtu.edu.cn}
% For a paper whose authors are all at the same institution,
% omit the following lines up until the closing ``}''.
% Additional authors and addresses can be added with ``\and'',
% just like the second author.
% To save space, use either the email address or home page, not both
}
\maketitle

\newcommand{\pb}[1]{{\leavevmode\color{red}[pb: #1]}}
\newcommand{\warn}[1]{{\leavevmode\color{red}[#1]}}

% {\color{red} 
% \begin{enumerate}
%     \item Shorten the intro.
%     \item Define region proxy: region feature + geometrics
%     \item Emphasize the computation primitive
%     \item Revisit consistency
% \end{enumerate}
% }

%%%%%%%%% ABSTRACT
\begin{abstract}

Typical vision backbones manipulate structured features. As a compromise, semantic segmentation has long been modeled as per-point prediction on dense regular grids. In this work, we present a novel and efficient modeling that starts from interpreting the image as a tessellation of learnable regions, each of which has flexible geometrics and carries homogeneous semantics. To model region-wise context, we exploit Transformer to encode regions in a sequence-to-sequence manner by applying multi-layer self-attention on the region embeddings, which serve as \textbf{proxies} of specific regions. Semantic segmentation is now carried out as per-region prediction on top of the encoded region embeddings using a single linear classifier, where a decoder is no longer needed. The proposed RegProxy model discards the common Cartesian feature layout and operates purely at region level. Hence, it exhibits the most competitive performance-efficiency trade-off compared with the conventional dense prediction methods. For example, on ADE20K, the small-sized RegProxy-S/16 outperforms the best CNN model using 25\% parameters and 4\% computation, while the largest RegProxy-L/16 achieves \SI{52.9}{mIoU} which outperforms the state-of-the-art by 2.1\% with fewer resources. Codes and models are available at \url{https://github.com/YiF-Zhang/RegionProxy}.

\end{abstract}

\vspace{-1em}

%%%%%%%%% BODY TEXT
\vspace{-0.1em}
\section{Introduction}
\vspace{-0.1em}

Semantic segmentation is one of the fundamental tasks in computer vision, and has been carried out using CNNs since the beginning of the deep learning era\cite{deng2009imagenet,krizhevsky2012imagenet,he2016deep,long2015fully}. However, CNN is not the out-of-the-box solution for semantic segmentation considering two of its natures: \begin{inparaenum}[1)]
    \item \emph{Limited context.} CNN lacks of abilities to capture long range dependencies for context modeling, which is essential for semantic segmentation.
    \item \emph{Coarse prediction.} Due to its hierarchical nature, CNN outputs coarse feature which is inadequate for dense labeling.
\end{inparaenum}
Fundamentally, the majority of semantic segmentation researches since FCN~\cite{long2015fully} have been centring on resolving these two issues. A great number of works have been proposed for better context modeling~\cite{zhao2017pyramid,zhang2018context,huang2019ccnet,yuan2020object} and fine-grained feature prediction~\cite{ronneberger2015u,chen2018encoder,wang2020deep}, which significantly advance the semantic segmentation research.

\begin{figure}
\centering
\includegraphics[width=1.0\linewidth]{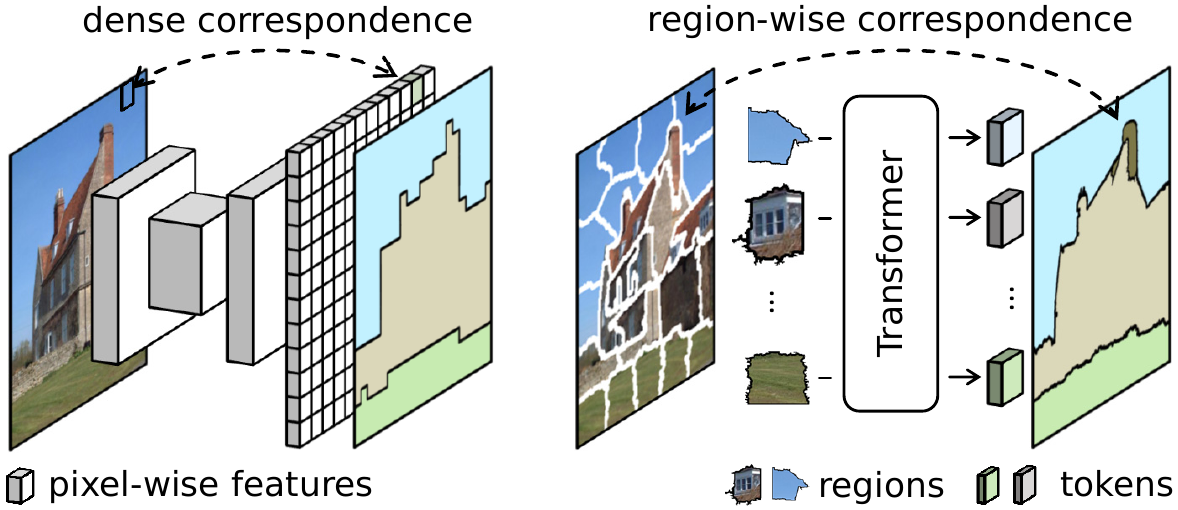}
\vspace*{-1.5em}
\caption{\textbf{Illustration of two different schemes for semantic segmentation.} \textbf{(Left)} Conventional encoder-decoder models establish dense correspondence between input and output on structured pixel-level grid, and segment image in a per-pixel prediction fashion. \textbf{(Right)} We propose to interpret the image as a tessellation of \emph{learnable} regions and represent it by region-level embeddings (i.e., \emph{proxies}) at an early stage, and segment image by \emph{per-region prediction} using sequence-to-sequence Transformer~\cite{vaswani2017attention, dosovitskiy2020image}.} 
\label{fig:illustration}
\vspace*{-1.5em}
\end{figure}

Currently, the Transformer architecture~\cite{vaswani2017attention} from natural language processing is introduced to the the vision community and has gained significant research interest. The Vision Transformer (ViT)~\cite{dosovitskiy2020image} partitions image into square patches and encodes their embeddings (i.e., \emph{tokens}) in a sequence-to-sequence manner using stacked self-attention layers. Some of its variants~\cite{liu2021swin,wang2021pyramid} adopt a hierarchical structure and restrict self-attention in local area for better scalability. 
% Vision transformers have shown their potential to function as general purpose vision backbones for many tasks. 
Recently, several semantic segmentation works~\cite{zheng2021rethinking,xie2021segformer,strudel2021segmenter} adopt vision Transformers as backbone and achieve impressive performances. They learn better context with the help of the inherent advantages of vision Transformers, namely the attention mechanism. However, in these models, the vision Transformer serves transparently as feature extractor which extracts 2D coarse features exactly as its convolutional counterpart does, while its main character, a sequence-to-sequence encoder, is not touched.

We revisit the aforementioned two issues: limited context and coarse prediction. While the former is a corollary of the local receptive field which can be alleviated by adopting Transformer architecture or previous CNN-based context modules, the latter is substantially induced by the inflexible \emph{regular} (Cartesian) layout of network features, as it does not follow the structure of real world semantics: with the large strides of typical vision backbones, it carries jumbly semantics in grid cells and brings difficulties for dense labeling. Hence, a ``decoder'' is required to produce fine-grained features. These facts imply that regular grids may not be the optimal feature arrangement for semantic segmentation.

In this work, we explore a novel modeling of semantic segmentation which we believe to be closer to its essence: we attempt to interpret image as a set of interrelated \emph{regions}, where the \emph{region} indicates a group of adjacent pixels with homogeneous semantics. As illustrated in Figure~\ref{fig:illustration}, we propose a simple \textbf{RegProxy} approach which \emph{learns} regions at an early stage, explicitly models inter-region relations using Transformer~\cite{vaswani2017attention}, and encodes regions in a sequence-to-sequence fashion. We design a novel mechanism to describe region geometrics and ensure the tessellation of the entire region set, which enables us to conduct semantic segmentation by \emph{per-region} prediction. The entire process is fully parameterized and differentiable which can be trained end-to-end efficiently. Here we present the major novelties and contributions of this work:
\begin{inparaenum}[1)]
    \item Instead of manipulating features on \emph{regular grids}, we operate on a set of region embeddings throughout the entire network. Each of the region embeddings serves as the feature representation of a specific learnable region, namely the \emph{region proxy}.
    \item Instead of using Transformer to extract structured feature, we dive into its essence as a sequence-to-sequence encoder, and use it to explicitly model inter-region relations.
    \item Instead of modeling semantic segmentation as per-pixel prediction using decoder, we segment images by directly predicting on the region embeddings using a linear classifier.
\end{inparaenum}

\begin{figure}[t]
\centering
\includegraphics[width=1.0\linewidth]{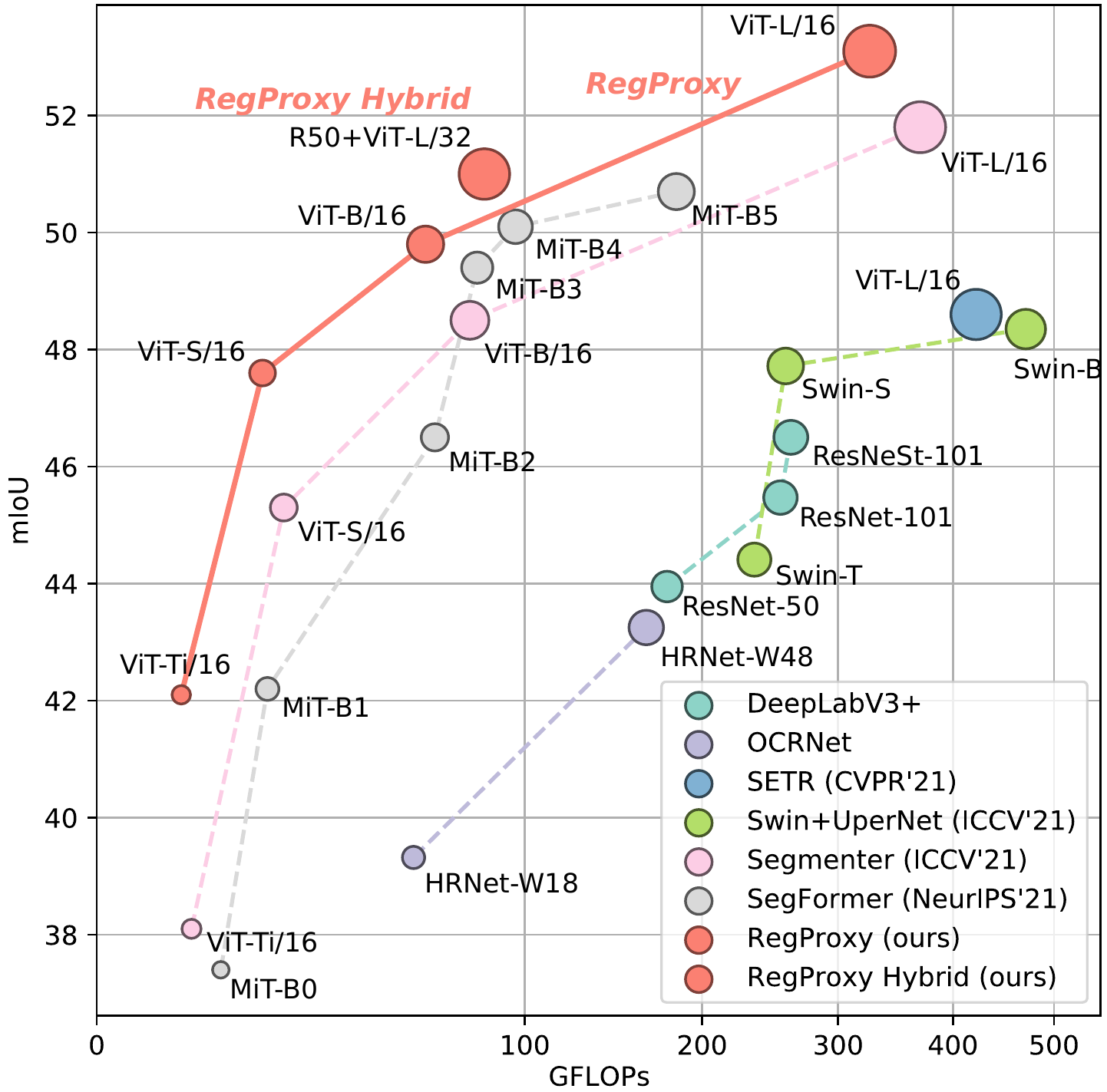} 
\vspace{-1.5em}
\caption{\textbf{Performance \textit{vs.} GFLOPs on ADE20K \texttt{val} split.} We report results {without} multi-scale inference. RegProxy shows a highly competitive performance-efficiency trade-off among the \emph{state-of-the-art} methods. Best viewed in color.}
\label{fig:perf-gflops-param}
\vspace{-1.5em}
\end{figure}

We build our model on bare ViTs~\cite{dosovitskiy2020image} for image classification by adding negligible overhead (\textasciitilde0.5\% parameters and GFLOPs), and consistently achieves the state-of-the-art performances throughout different model sizes. Extensive experiments show the competitive performance-efficiency trade-off of RegProxy under various model capacities on multiple datasets. One may peek the results in Figure~\ref{fig:perf-gflops-param}.

\vspace{-0.1em}
\section{Related Work}
\label{sec:related}
\vspace{-0.1em}

\paragraph{Vision Transformer}

Transformer~\cite{vaswani2017attention} was firstly introduced in machine translation and is currently the de-facto standard of most natural language processing (NLP) tasks. The proposed attention mechanism has also inspired a number of works in computer vision~\cite{wang2018non, hu2018squeeze, woo2018cbam}. Notably, the attention works particularly well in semantic segmentation~\cite{fu2019dual, huang2019ccnet, li2019expectation, zhu2019asymmetric} where contextual information is crucial.

Recently, Dosovitskiy et al.~\cite{dosovitskiy2020image} proposes the Vision Transformer (ViT) which directly inherits the Transformer~\cite{vaswani2017attention} architecture from NLP and works as a standalone model. ViT is gaining significant research interest and a number of improved models have been proposed. Generally, the architecture of vision Transformers can be sequential or hierarchical. Sequential models (including the original ViT~\cite{dosovitskiy2020image}) partition image into patches and encode inputs in a sequence-to-sequence fashion by computing the global self-attention. DeiT~\cite{touvron2021training} successfully trains ViTs on ImageNet-1k~\cite{deng2009imagenet} with the help of strong data augmentations and knowledge distillation. CaiT~\cite{touvron2021going} proposes LayerScale technique and later class tokens for effective training of vision Transformers at depth. XCiT~\cite{el2021xcit} proposes cross-covariance attention that operates on feature dimension to build more scalable vision Transformers. Hierarchical models~\cite{liu2021swin, wang2021pyramid} borrow some of the image-specific inductive bias from CNNs, such as translation equivariance and 2D neighborhood structure on regular grid. They compute attentions within local windows and produce hierarchical 2D features. Swin Transformer~\cite{liu2021swin} is one of the most successful hierarchical vision Transformers. With the shifted window design, it brings better efficiency while allowing cross-window connections for better feature extraction.

\paragraph{Semantic Segmentation} Semantic segmentation has been modeled as dense prediction since the emergence of fully conventional architectures. FCN~\cite{long2015fully} lays the foundation of modern semantic segmentation models, which is the first to adopt fully convolutional neural networks to segment images with arbitrary scales in an end-to-end manner. The following works inherit the fully convolutional design and focus on better context modeling. The efforts have been put on one or more aspects that significantly improve semantic segmentation performance:
\begin{inparaenum}[1)]
    \item Enlarge receptive field using larger kernels or dilated convolution~\cite{chen2014semantic,chen2017deeplab,yu2015multi,peng2017large};
    \item Integrate multi-scale features~\cite{zhao2017pyramid,chen2017rethinking,chen2018encoder,kirillov2019panoptic};
    \item Refine contextual information~\cite{zhang2018context,yuan2020object,lin2017refinenet};
    \item Utilize attention~\cite{fu2019dual, huang2019ccnet, li2019expectation, zhu2019asymmetric};
    \item Search or design designated backbones~\cite{wang2020deep,li2020learning,nekrasov2019fast,ranftl2021vision}.
\end{inparaenum}

Recently, several works exploit Transformer to conduct semantic segmentation. SETR~\cite{zheng2021rethinking} is the first to introduce vision Transformer backbones into semantic segmentation. Segmenter~\cite{strudel2021segmenter} exploits Transformer to predict per-class masks. SegFormer~\cite{xie2021segformer} and DPT~\cite{ranftl2021vision} propose hierarchical vision Transformer backbones specifically designed for dense prediction tasks. MaskFormer~\cite{cheng2021per} uses the Transformer Decoder~\cite{vaswani2017attention} to query classes with their masks from a conventional encoder-decoder model. We emphasize that our work has significant difference with these approaches: we use \emph{bare} Transformer to model \emph{inter-region} relations instead of using it as replaceable or extra modules of existing dense-prediction architectures. 
% {\color{red} [Review this claim.]}

\begin{table}
    \footnotesize
    \centering
    \begin{adjustbox}{max width=1.0\linewidth}
    \begin{threeparttable}
    \tablestyle{0.6\tabcolsep}{1.15}
    \begin{tabular}{ll|cc|c@{\hspace{0.5\tabcolsep}}l|c@{\hspace{0.5\tabcolsep}}l}
    backbone  & method   & \#params.       & FLOPs           & \multicolumn{2}{c|}{ADE20K}                                   & \multicolumn{2}{c}{Cityscapes}                                \\ \shline
    ViT-Ti/16 & Baseline & \phantom{0}5.7M & \phantom{0}3.8G & 39.0          & \smallscript{/ 37.8}                          & 72.3          & \smallscript{/ 68.1}                          \\
              & Ours     & -               & -               & \textbf{40.9} & {\color[HTML]{009901} \cgap{+}{\textbf{1.9}}} & \textbf{74.1} & {\color[HTML]{009901} \cgap{+}{\textbf{1.8}}} \\ \hline
    ViT-S/16  & Baseline & 22.0M           & 14.9G           & 45.4          & \smallscript{/ 44.2}                          & \textbf{76.1} & \smallscript{/ 71.8}                          \\
              & Ours     & -               & -               & \textbf{46.0} & {\color[HTML]{009901} \cgap{+}{\textbf{0.6}}} & 75.9          & {\color[HTML]{9A0000} \cgap{-}{0.2}}          \\ \hline
    ViT-B/16  & Baseline & 86.6M           & 58.8G           & 47.1          & \smallscript{/ 45.6}                          & \textbf{78.5} & \smallscript{/ 75.1}                          \\
              & Ours     & -               & -               & \textbf{47.3} & {\color[HTML]{009901} \cgap{+}{\textbf{0.2}}} & 77.3          & {\color[HTML]{9A0000} \cgap{-}{1.2}}         
    \end{tabular}
    \end{threeparttable}
    \end{adjustbox}
    \vspace{-0.5em}
    \caption{\textbf{Pilot experiment results.} We report baseline results with/{\color[HTML]{9B9B9B}without} the bilinear upsampling of the final logits map. The baseline uses the original patch embedding~\cite{dosovitskiy2020image} and works as a typical segmentation model with stride 16, while our prototype model embeds non-parametric regions (i.e., superpixels).\label{tab:pilot}}
    \vspace{-1.5em}
    \end{table}

\vspace{-0.1em}
\section{\mbox{Pilot Study of Region Proxy}}
\label{sec:pilot}
\vspace{-0.1em}

Before introducing our main model, we first present a naive instantiation of region proxy for a proof of concept. We revisit the classical superpixel segmentation~\cite{ren2003learning}. Superpixels are an over-segmentation of images that is formed by grouping perceptually similar pixels together based on low-level information (e.g., colors). It provides a low/mid-level image representation for high-level vision tasks including semantic segmentation~\cite{gould2008multi,sharma2014recursive,gadde2016superpixel}. In our pilot study, we adopt widely used SLIC~\cite{achanta2012slic} method to generate non-parametric regions to validate our region proxy notion.

\paragraph{Regions as Superpixels}
% \label{sec:spt}

Essentially, we attempt to batch the pixel labeling by classifying superpixels. This implies that nearby pixels with similar low-level properties should be semantically homogeneous. This assumption will finally be analyzed in Section~\ref{sec:analysis} along with our main model introduced in Section~\ref{sec:method}. At the moment, we build our prototype model under this premise.

We modify the vanilla ViT~\cite{dosovitskiy2020image} which encodes patches, making it feasible to encode superpixels: $N$ superpixels $\{\mathbf{s}_i\}$ are generated from input image $\mathbf{I}\in \mathbb{R}^{H\times W\times 3}$. The irregular superpixels are cropped by their enclosing bounding boxes and resized to fixed size patches $\{\mathbf{x}_i\}$ where $\mathbf{x}_i\in \mathbb{R}^{P\times P\times 3}$, which is implemented using ROIAlign~\cite{he2017mask}. 
%  We also utilize the shape information by embedding their binary masks $\{\mathbf{m}_i\}$ within the cropped bounding, where $\mathbf{m}_i\in\mathbb{R}^{P\times P\times 1}$. 
 We embed $\mathbf{x}_i$ using the exact linear patch embedding in the vanilla ViT, producing $N$ tokens with dimension $D$, which are then encoded in a sequence-to-sequence manner. Instead of per-pixel prediction, We directly classify regions by applying a linear classifier on every tokens. For supervision, a soft label $\hat{\mathbf{y}}_i$ is applied on the $i$-th token. We have
\begin{equation}
    \label{eq:class-hist}
    \hat{\mathbf{y}}_i[c] = \frac{\vert\{\mathbf{p}\in \mathbf{s}_i \mid \hat{y}(\mathbf{p}) = c\}\vert}{\vert \mathbf{s}_i \vert}, c\in\{0,1,\dots,K-1\}
\end{equation}
where $\hat{y}(\mathbf{p})$ is the label of the pixel $\mathbf{p}$, and $K$ is the number of classes. Namely, $\hat{\mathbf{y}}_i$ is the category histogram of pixels inside superpixel $\mathbf{s}_i$.

\paragraph{Experiment} To set up the baseline, we apply the same linear classifier to the output tokens of the vanilla ViT. Hence the only difference between the baseline and our prototype model is the computation primitive: whether to be a square patch or a region. The training protocol is identical to our main experiments which is described in Section~\ref{sec:experiments}. In Table~\ref{tab:pilot}, We report results of three model variants ViT-\{Ti/16, S/16, B/16\} on ADE20K and Cityscapes. We observe evidential performance gains on smaller models, which vanish as the model getting larger. We hypothesize that the introduced low-level priori benefits small models, however, its intrinsic noise (i.e., inaccurate superpixels) becomes the bottleneck for larger models. We also notice the importance of the final bilinear interpolation as a standard practice: without it, the baseline model conducts patch classification which is a degeneration of the superpixel model, and yields worse performance.

\vspace{-0.1em}
\section{Method}
\label{sec:method}
\vspace{-0.1em}

\begin{figure*}
\vspace{-1em}
\centering
\includegraphics[width=1.0\linewidth]{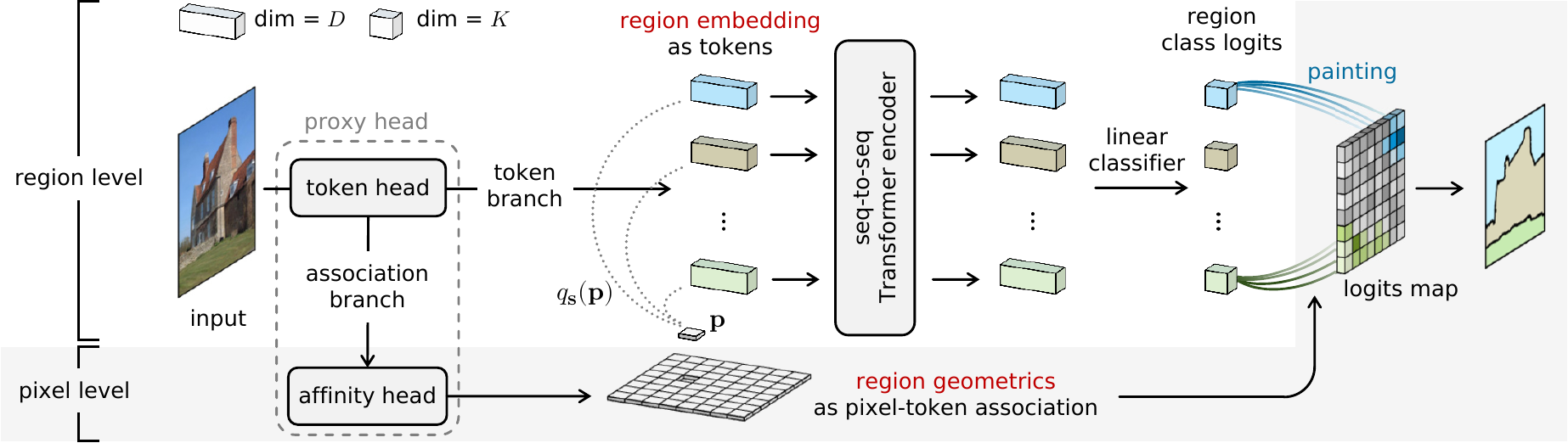}   
\vspace{-1.5em}
\caption{\textbf{Overview of our RegProxy approach.} The sequence-to-sequence Transformer encoder computes on {\color{Maroon}region embeddings} in the form of tokens, which serve as \emph{proxies} of specific regions whose {\color{Maroon}geometrics} are described by the class-agnostic pixel-token association. Notably, we model global context completely at region-level without any typical ``feature map''. The region embedding and its geometrical description are jointly learned using the \emph{proxy head}. A single linear classifier is adopted for \emph{per-region} prediction. The region class logits are simply ``{\color{RoyalBlue}painted}'' to the output plane according the corresponding region geometrics to yield final segmentation result.}
\label{fig:main}
\vspace{-1em}
\end{figure*}

The aforementioned pilot experiments motivate us to explore for better instantiation of the region proxy modeling. In this section, we introduce the RegProxy model which computes on region embeddings (as tokens) that serve as computational proxies of specific \emph{learnable} regions, and exploit Transformer~\cite{vaswani2017attention, dosovitskiy2020image} to model region-wise context. 

We first provide basic information about the Transformer for region context modeling in Section~\ref{sec:transformer-region-encoder}. We introduce the learning of region geometrics and the embedding of region features in Section~\ref{sec:pixel-token-association}. Finally, we review RegProxy model from the system perspective and describe the training/inference as per-region prediction in Section~\ref{sec:per-retion-prediction}.

\subsection{Transformer as Region Encoder}
\label{sec:transformer-region-encoder}

Transformer is a type of sequence-to-sequence model, which applies multi-layer self-attention on its computation primitives, i.e. \emph{tokens}. In this work, we take full advantage of the Transformer architecture to learn global context for image by directly modeling inter-region relations using self-attention. While the region learning and embedding will be discussed in Section~\ref{sec:pixel-token-association}, in this section, we briefly introduce the principle of the Transformer encoder.

Supposing $N$ region embeddings has been produced, we write the Transformer input as a sequence of tokens:
% \pb{something wrong}:
\begin{equation}
    \mathbf{X}_0 = \begin{bmatrix}
        \mathbf{x}_\texttt{cls}^0, \mathbf{x}_0^0, \mathbf{x}_1^0, \dots, \mathbf{x}_{N-1}^0
    \end{bmatrix}^{\top}\in \mathbb{R}^{(N+1)\times D},
\end{equation}
where $D$ is the embedding dimension and the class token $\mathbf{x}_\texttt{cls}$ serves as a whole-image representation in image recognition tasks. A Transformer encoder is composed of $L$ stacked layers, each of which consists of a multi-head self-attention (MSA) block followed by a multi-layer perceptron (MLP) block with two linear projection layers. Layer normalization\cite{ba2016layer} and residual connections are applied for both MSA and MLP blocks. Mathematically, we can write the computation of layer $l$ as
\begin{equation}
\begin{split}
    \mathbf{A}_l &= \operatorname{MSA}\left(\operatorname{LN}\left(\mathbf{X}_{l-1}\right)\right) + \mathbf{X}_{l-1},\\
    \mathbf{Z}_l &= \operatorname{MLP}\left(\operatorname{LN}\left(\mathbf{A}_{l-1}\right)\right) + \mathbf{A}_{l-1},\\
\end{split}
\end{equation}
where the output $\mathbf{Z}_l$ is used as input of the next layer.
% {\color{red} Write details here or guide the audience to appendix.}

\begin{figure}
    \centering
    \begin{subfigure}[b]{0.64\linewidth}
        \centering
        \includegraphics[width=1.0\textwidth]{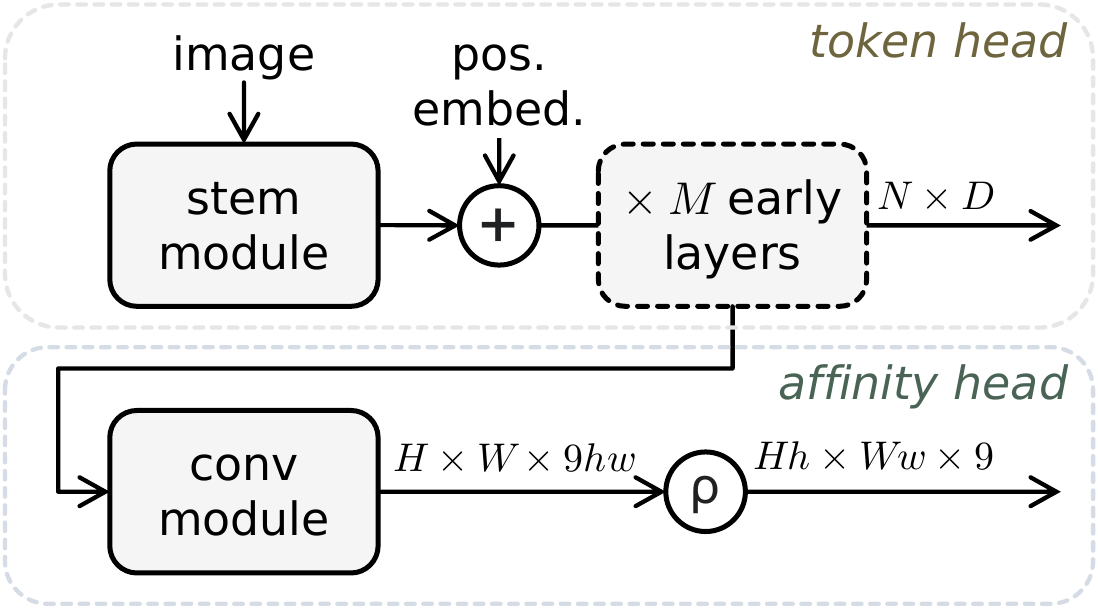}   
        \caption{}
        \label{fig:stem-affinity-head}
    \end{subfigure}
    \hspace{0.02\linewidth}
    \begin{subfigure}[b]{0.24\linewidth}
        \centering
        \includegraphics[width=1.0\textwidth]{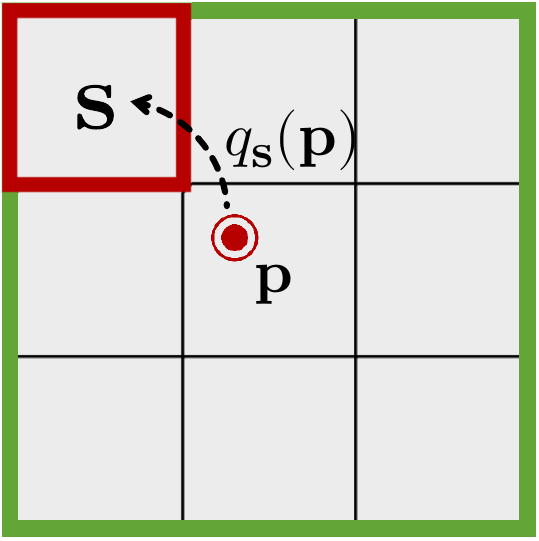}
        \caption{}
        \label{fig:pixel-token-association}
    \end{subfigure}
    \vspace{-0.5em}
    \caption{\textbf{More details of the RegProxy model.} (a) Illustration of the proxy head, where \texttt{+} stands for element-wise addition and $\mathgtt{r}$ stands for reshape and rearrange of dimensions. (b) Describing region geometrics by local pixel-region association.}
    \vspace{-1.5em}
    \end{figure}

The Vision Transformer (ViT) inherits the exact architecture from the Transformer in NLP~\cite{vaswani2017attention}, accompanied with a stem module designed for vision tasks. In this work, we follow the standard definition of the ViT models in \cite{dosovitskiy2020image, touvron2021training}. Our RegProxy-\{Ti/16, S/16, B/16, L/16\} models are based on vanilla ViTs which use patch embedding stems, while our RegProxy-\{R26+S/32, R50+L/32\} models are based on hybrid ViTs which use CNN stems. These ViTs vary from total strides and numbers of layers, embed dimensions and attention heads. The detailed specifications can be found in Table~\ref{tab:vit-spec} and are described in Section~\ref{sec:experiments}.

\subsection{Learning Region Proxy}
\label{sec:pixel-token-association}

As depicted in Figure~\ref{fig:main}, our \emph{region proxy} method comes with two important components: the region feature and its geometrics. In this section, we introduce the region feature embedding and its geometrical description.

\paragraph{Describing Region Geometrics} Learning and describing region geometrics is not quite straightforward due to the irregular shape and various scales. An intuitive idea is to predict a binary mask for each region, either to be image-sized or within a local area. However, we find neither of the approaches practical for our sequence-to-sequence modeling. Predicting a full-sized mask for each token will be computationally expensive, while predicting small sized local masks cannot ensure the tessellation of regions, consequently causing overlapping and non-predicted areas.

To this end, we propose a novel mechanism to describe region geometrics by pixel-to-token association. We start from an initial $H\times W$ grid where $H\times W=N$. Each token lays on a single cell which serves as a ``seed'' of its corresponding region $\mathbf{s}$. Note the cell itself is merely a token location indicator and has nothing to do with the actual region geometrics. We build the pixel-to-token association by assigning each pixel $\mathbf{p}=(u,v)$ to region $\mathbf{s}$ with a probability $q_\mathbf{s}(\mathbf{p})$. It is unnecessary to apply this association globally. Instead, we associate pixel $\mathbf{p}$ only with tokens that lays in its neighborhood $\mathcal{N}_\mathbf{p}$ which satisfies
\begin{equation}
\label{eq:prob-norm}
    \sum_{\mathbf{s}\in\mathcal{N}_{\mathbf{p}}}q_\mathbf{s}(\mathbf{p}) = 1
\end{equation}
This is illustrated in Figure~\ref{fig:pixel-token-association}, as the pixel $\mathbf{p}$ is being assigned to one of the 9 region seeds (marked with red box) with in its surrounding neighborhood (in green box). Mathematically, we can write the association as a class-agnostic affinity map $\mathbf{Q}\in\mathbb{R}^{(Hh)\times (Ww)\times \vert\mathcal{N}_p\vert}$ at pixel-level which describe the actual geometrics of all regions, where $(Hh)\times(Ww)$ matches the size of the output segmentation map and $(h,w)$ is the relative stride of the initial token grid. Empirically, we set the neighborhood size to $3\times 3$ and $\vert\mathcal{N}_p\vert=9$, which works well through-out all model sizes.

For an intuitive interpretation, the region is represented by a ``core'' (i.e., token's location on the initial $H\times W$ cell) and pixels surround it by probabilities. With the constraint of Eq.~\ref{eq:prob-norm}, the entire set of probabilistic regions forms a tessellation that covers the image plane with no overlaps and no gaps. Meanwhile, the regions learn to be highly semantically homogeneous even without explicit regularization, which will be discussed in Section~\ref{sec:analysis}. These facts enable us to segment image in a per-region prediction fashion.

\paragraph{Embedding Regions} We jointly embed region features and learn their geometrical description $\mathbf{Q}$ using the \emph{proxy head} depicted in Figure~\ref{fig:stem-affinity-head}. To embed region features into tokens, we use a small part of the ViT backbone as \emph{token head}. We take the exact stem module in \cite{dosovitskiy2020image} along with learnable position embeddings, which produces $N$ tokens of dimension $D$ where $N=H\times W$. The stem module can be linear patch projection for vanilla ViT models, or CNNs for hybrid models (see Section~\ref{sec:experiments}).
%  For our hybrid models RegProxy-\{R26+S/32, R50+L/32\}, we directly use the output embeddings of stem module as the input tokens of transformer encoder. 
For our vanilla ViT models, we further apply the first $M$ Transformer layers for better feature extraction.

To predict region geometrical description $\mathbf{Q}$, we design a lightweight \emph{affinity head} that is highly integrated with the ViT backbone, bringing negligible extra parameters and computations. As illustrated in Figure~\ref{fig:stem-affinity-head}, we predict the affinity map $\mathbf{Q}$ directly using features from the token head. Specifically, the token feature $\mathbf{T}\in \mathbb{R}^{N\times D}$ is reshaped to $\mathbf{T}^\prime\in \mathbb{R}^{H\times W\times D}$ according to its initial layout, and then fed into a convolution module. The convolution module fuses local region-wise information and produces $H\times W\times (9hw)$ feature map, as we predict affinity vectors for all $hw$ pixels inside each cell in batches. The compact map is then ``unpacked'' to the pixel-level format, which finally yields the affinity map $\mathbf{Q}\in\mathbb{R}^{(Hh)\times (Ww)\times 9}$.

In details, the convolution module consists of one $3\times 3$ depth-wise conv layer, followed by a $1\times 1$ conv layer. The affinity map is activated by a Softmax to produce normalized probabilities. Notably, we discover some interesting facts about choices of the number of early layers $M$ for affinity prediction, which will be discussed in Section~\ref{sec:analysis}.

\subsection{Segmentation by Per-Region Prediction}
\label{sec:per-retion-prediction}

By our assumption, region embedding carries homogeneous semantics throughout the Transformer network. This enables us to directly predict on regions instead of pixels as opposed to conventional FCN-style models~\cite{long2015fully,chen2018encoder,liu2021swin,xie2021segformer}, hence significantly reduces the number of predictions. As depicted in Figure~\ref{fig:main}, we apply a \textbf{linear} classifier on the last layer output $\mathbf{X}_L$ of the Transformer backbone, which we find sufficient to yield strong performance. The linear classifier produces class logits for all tokens except $\mathbf{x}_\texttt{cls}$:
\begin{equation}
    \mathbf{Y} = \begin{bmatrix}
        \mathbf{y}_0, \mathbf{y}_1, \dots, \mathbf{y}_{N-1}
    \end{bmatrix}^\top\in \mathbb{R}^{N\times K},
\end{equation}
where $K$ is number of classes.

We can easily attain the pixel-level segmentation result using the region geometrical description $\mathbf{Q}$ introduced in Section \ref{sec:pixel-token-association}. We simply ``paint'' the token logits to the corresponding regions as illustrated in Figure~\ref{fig:main}. In practice, we paint in a per-pixel manner for efficient implementation. For pixel $\mathbf{p}=(u,v)$, its class logits is calculated by
\begin{equation}
    \mathbf{Y}^\prime[u,v] = \sum_{\mathbf{s}\in\mathcal{N}_\mathbf{p}} \mathbf{y}(\mathbf{s})\cdot q_\mathbf{s}(\mathbf{p}),
\end{equation}
where $\mathbf{Y}^\prime\in \mathbb{R}^{(Hh)\times(Ww)\times K}$ is the output logits map, $\mathbf{y}(\mathbf{s})$ is the class logits of the token corresponding to region $\mathbf{s}$. In practice, we set $(h, w) = (4, 4)$, yielding $\times 4$ stride logits map for RegProxy-$x$/16 models, and $\times 8$ stride logits map for RegProxy-$x$/32 models. Our model is trained end-to-end using cross-entropy loss \emph{without} class balance or hard example mining. The output logits map is upsampled to ground-truth/input size for supervision during training and prediction during inference following the standard practice.
 
\vspace{-0.1em}
\section{Experiments}
\label{sec:experiments}
\vspace{-0.1em}

\paragraph{Datasets and Metrics} We study RegProxy on three public datasets: ADE20K~\cite{zhou2017scene}, Cityscapes~\cite{cordts2016cityscapes} and Pascal Context~\cite{mottaghi2014role}. ADE20K is a challenging scene parsing dataset with 150 classes, which contains \num{20210} images for training and \num{2000} images for validation. Cityscapes is a high-resolution street scene dataset with 19 classes. We train on the \num{5000} fine annotated training images and test on \num{500} validation images. Pascal Context contains \num{4996} training images and \num{5104} validation images with 60 classes (including one background class). We report Intersection over Union averaged over all classes (mIoU).

\paragraph{ViT Backbones} We build our model on the original ViTs as defined in~\cite{dosovitskiy2020image,steiner2021train}. We conduct extensive experiments on six variants: vanilla ViTs with four specs (tiny, small, base and large) and hybrid ViTs with two specs (small, large). Vanilla ViTs use \emph{patch embedding} as stem which partitions an image into $P\times P$ patches, while each patch is flattened and linearly projected to an embedding with $D$ dimension. For hybrid ViTs, the projection is applied to the features extracted by a residual convolutional network~\cite{he2016deep}, hence the patch size is equivalent to CNN's stride $S$. We set $P=16$ for vanilla ViTs and $S=32$ for hybrid ones. Namely, the initial $H\times W$ token grid is of strides $16$ and $32$ respectively. Larger stride generally impairs performance while requiring less computation. The detailed configurations can be found in Table~\ref{tab:vit-spec}. We use weights pre-trained on ImageNet-21k~\cite{steiner2021train} following recent works~\cite{zheng2021rethinking, strudel2021segmenter, liu2021swin}.

\begin{table}[t]
\vspace{-1em}
\small
\centering
\begin{adjustbox}{max width=1.0\linewidth}
\tablestyle{0.6\tabcolsep}{1.2}\setlength\tabcolsep{0.5em}
\begin{tabular}{l|c@{\hspace{0.7\tabcolsep}}c@{\hspace{0.7\tabcolsep}}|c@{\hspace{0.7\tabcolsep}}c@{\hspace{0.7\tabcolsep}}c@{\hspace{0.7\tabcolsep}}|r}
backbone & stride & stem               & \#layers & dim  & \#heads & \#params       \\ \shline
Ti/16    & 16     & patch embed.       & 12       & 192  & 3       & \phantom{00}6M \\
S/16     & 16     & patch embed.       & 12       & 384  & 6       & \phantom{0}22M \\
B/16     & 16     & patch embed.       & 12       & 768  & 12      & \phantom{0}86M \\
L/16     & 16     & patch embed.       & 24       & 1024 & 16      & 307M           \\ \hline
R26+S/32 & 32     & ResNet[2, 2, 2, 2] & 12       & 384  & 6       & \phantom{0}37M \\
R50+L/32 & 32     & ResNet[3, 4, 6, 3] & 24       & 1024 & 16      & 330M          
\end{tabular}
\end{adjustbox}
\vspace{-0.5em}
\caption{\textbf{Configurations of ViT variants.}\label{tab:vit-spec} (Top) Vanilla ViTs use patch embedding. (Bottom) Hybrid ViTs use ResNet~\cite{he2016deep} stem.}
\vspace{-1.5em}
\end{table}

\paragraph{Implementation Details}

We implement our method using the public codebase \texttt{mmsegmentation}~\cite{mmseg2020}. We introduce minimum changes to its default settings that are widely used by the community. For training, we use input sizes of $512\times 512$, $768\times 768$ and $480\times 480$ for ADE20K, Cityscapes and Pascal Context, respectively. We train our ``Large'' model on ADE20K using a $640\times 640$ crop following \cite{liu2021swin,xie2021segformer,strudel2021segmenter}. We use AdamW~\cite{loshchilov2017decoupled} optimizer with an initial learning rate of $6\times 10^{-5}$, a weight decay of $10^{-2}$ and a ``poly'' learning rate scheduler~\cite{chen2017deeplab} with $\mathit{power}=1.0$ following \cite{liu2021swin,xie2021segformer,el2021xcit}. We use a batch size of 16 and train for 160k, 80k, 40k iterations on ADE20K, Cityscapes and Pascal Context, respectively. We keep data augmentations and all other training settings identical to default settings in~\cite{mmseg2020}. Training tricks such as hard example mining, auxiliary losses or class balance loss are \emph{not} included.

For testing, we use the sliding window mode with window sizes matching the crop sizes for training. We use default window strides in \cite{mmseg2020}. We report both single scale results and multi-scale + flipping results with scaling factors of $\{0.5, 0.75, 1.0, 1.25, 1.5, 1.75\}$.

\subsection{Main Results}

\begin{table}[t]
\vspace{-1em}
\footnotesize
\centering
\begin{adjustbox}{max width=1.0\linewidth}
\begin{threeparttable}
\tablestyle{0.6\tabcolsep}{1.2}
\begin{tabular}{l@{\hspace{0.4\tabcolsep}}|c<{\hspace{-\tabcolsep}}>{\hspace{-\tabcolsep}\,}c@{\hspace{0.4\tabcolsep}}|c<{\hspace{-\tabcolsep}\,}>{\hspace{-\tabcolsep}\,}c@{\hspace{1.0\tabcolsep}}c<{\hspace{-\tabcolsep}}>{\hspace{-\tabcolsep}\,}c@{\hspace{0.0\tabcolsep}}|c<{\hspace{-\tabcolsep}\,}>{\hspace{-\tabcolsep}\,}c@{\hspace{1.0\tabcolsep}}c<{\hspace{-\tabcolsep}}>{\hspace{-\tabcolsep}\,}c@{\hspace{0.0\tabcolsep}}}
method                                                 & FLOPs                                   & \#params.                               & \multicolumn{4}{c|}{ADE20K (SS/MS)}                                                                                                                       & \multicolumn{4}{c}{Cityscapes (SS/MS)}                                                                                                                 \\ \shline
ViT-Ti/16                                              & \phantom{00}3.8G                        & \phantom{00}5.7M                        & 39.0                        &                                               & 39.8                        &                                               & 72.3                        &                                               & 74.1                     &                                               \\
\, +Mask.T                                    & \phantom{0}+1.0G                        & \phantom{0}+1.0M                        & 38.1                        & {\color[HTML]{9A0000} \cgap{-}{0.9}}          & 38.8                        & {\color[HTML]{9A0000} \cgap{-}{1.0}}          & -                           &                                               & -                        &                                               \\
\rowcolor{table-bg}
\textbf{\, +Ours}                                      & \phantom{0}+0.1G                        & \phantom{0}+0.1M                        & \textbf{42.1}               & {\color[HTML]{009901} \cgap{+}{\textbf{3.1}}} & \textbf{43.1}               & {\color[HTML]{009901} \cgap{+}{\textbf{3.3}}} & 76.5                        & {\color[HTML]{009901} \cgap{+}{4.2}}          & 77.7                     & {\color[HTML]{009901} \cgap{+}{3.6}}          \\ \hline
ViT-S/16                                               & \phantom{0}14.9G                        & \phantom{0}22.0M                        & 45.4                        &                                               & 45.9                        &                                               & 76.1                        &                                               & 78.0                     &                                               \\
\, +Mask.T                                             & \phantom{0}+4.2G                        & \phantom{0}+4.1M                        & 45.3                        & {\color[HTML]{9A0000} \cgap{-}{0.1}}          & 46.9                        & \cgap{+}{1.0}                                 & -                           &                                               & -                        &                                               \\
\rowcolor{table-bg}
\textbf{\, +Ours}                                      & \phantom{0}+0.2G                        & \phantom{0}+0.2M                        & \textbf{47.6}               & {\color[HTML]{009901} \cgap{+}{\textbf{2.2}}} & \textbf{48.4}               & {\color[HTML]{009901} \cgap{+}{\textbf{2.5}}} & 79.8                        & {\color[HTML]{009901} \cgap{+}{3.7}}          & 81.5                     & {\color[HTML]{009901} \cgap{+}{3.5}}          \\ \hline
ViT-B/16                                               & \phantom{0}58.8G                        & \phantom{0}86.6M                        & 47.1                        &                                               & 48.1                        &                                               & 78.5                        &                                               & 80.5                     &                                               \\
\, +UperNet                                            & \phantom{.}+336G                        & +57.6M                                  & 47.9                        & \cgap{+}{0.8}                                 & 49.5                        & \cgap{+}{1.4}                                 & 79.6                        & \cgap{+}{1.1}                                 & 80.9                     & \cgap{+}{0.4}                                 \\
\, +Mask.T                                             & +17.1G                                  & +16.0M                                  & 48.7                        & \cgap{+}{1.6}                                 & 50.1                        & {\color[HTML]{009901} \cgap{+}{2.0}}          & -                           &                                               & 80.6                     & \cgap{+}{0.1}                                 \\
\rowcolor{table-bg}
\textbf{\, +Ours}                                      & \phantom{0}+0.7G                        & \phantom{0}+0.7M                        & \textbf{49.8}               & {\color[HTML]{009901} \cgap{+}{\textbf{2.7}}} & \textbf{50.5}               & {\color[HTML]{009901} \cgap{+}{\textbf{2.4}}} & \textbf{80.9}               & {\color[HTML]{009901} \cgap{+}{\textbf{2.4}}} & \textbf{82.2}            & \cgap{+}{\textbf{1.7}}                        \\ \hline
ViT-L/16                                               & 325.0G                                  & 304.3M                                  & 50.7                        &                                               & 51.8                        &                                               & 78.4                        &                                               & 80.7                     &                                               \\
\, +Mask.T                                             & +44.5G                                  & +28.5M                                  & 51.8                        & \cgap{+}{1.1}                                 & \textbf{53.6}               & \cgap{+}{\textbf{1.8}}                        & -                           &                                               & 81.3                     & \cgap{+}{0.6}                                 \\
\rowcolor{table-bg}
\textbf{\, +Ours}                                      & \phantom{0}+0.9G                        & \phantom{0}+1.8M                        & \textbf{52.9}               & {\color[HTML]{009901} \cgap{+}{\textbf{2.2}}} & 53.4                        & \cgap{+}{1.6}                                 & \textbf{81.4}               & {\color[HTML]{009901} \cgap{+}{\textbf{3.0}}} & \textbf{82.7}            & {\color[HTML]{009901} \cgap{+}{\textbf{2.0}}} \\ \hline\hline
{\color[HTML]{9B9B9B} SETR~\cite{zheng2021rethinking}} & {\color[HTML]{9B9B9B} 325.1G}           & {\color[HTML]{9B9B9B} 305.6M}           & {\color[HTML]{9B9B9B} 48.1} & {\color[HTML]{9B9B9B} }                       & {\color[HTML]{9B9B9B} 48.8} & {\color[HTML]{9B9B9B} }                       & {\color[HTML]{9B9B9B} 77.9} & {\color[HTML]{9B9B9B} }                       & {\color[HTML]{9B9B9B} -} & {\color[HTML]{9B9B9B} }                       \\
{\color[HTML]{9B9B9B} \, +MLA}                         & {\color[HTML]{9B9B9B} \phantom{0}+8.7G} & {\color[HTML]{9B9B9B} \phantom{0}+4.0M} & {\color[HTML]{9B9B9B} 48.6} & {\color[HTML]{9B9B9B} \cgap{+}{0.5}}          & {\color[HTML]{9B9B9B} 50.3} & {\color[HTML]{9B9B9B} \cgap{+}{1.5}}          & {\color[HTML]{9B9B9B} 77.2} & {\color[HTML]{9B9B9B} \cgap{-}{0.7}}          & {\color[HTML]{9B9B9B} -} & {\color[HTML]{9B9B9B} }                       \\
{\color[HTML]{9B9B9B} \, +PUP}                         & {\color[HTML]{9B9B9B} +97.5G}           & {\color[HTML]{9B9B9B} +11.7M}           & {\color[HTML]{9B9B9B} 48.6} & {\color[HTML]{9B9B9B} \cgap{+}{0.5}}          & {\color[HTML]{9B9B9B} 50.1} & {\color[HTML]{9B9B9B} \cgap{+}{1.3}}          & {\color[HTML]{9B9B9B} 79.3} & {\color[HTML]{9B9B9B} \cgap{+}{1.4}}          & {\color[HTML]{9B9B9B} -} & {\color[HTML]{9B9B9B} }                      
\end{tabular}
\begin{tablenotes}
\footnotesize
    \item[*] In green are the gaps of at least {\color[HTML]{009901} +\textbf{2.0}} mIoU.
\end{tablenotes}
\end{threeparttable}
\end{adjustbox}
\vspace{-0.75em}
\caption{\textbf{Compare different integrations of vision Transformer.}\label{tab:main} We report the results of the baseline, the state-of-the-art Mask Transformer from Segmenter~\cite{strudel2021segmenter}, UperNet~\cite{xiao2018unified} and our RegProxy. We also report numbers from SETR~\cite{zheng2021rethinking} for reference.}
\vspace{-1.5em}
\end{table}

\paragraph{Baselines}
To setup the baseline, we simply append a linear classifier to the bare ViTs to produce per-patch prediction $\mathbf{Y}\in\mathbb{R}^{N\times K}$. Then the patch-wise logits are reshaped to their original 2D layout $\mathbf{Y}^\prime\in\mathbb{R}^{H\times W\times K}$ and upsampled to image size for training and inference following the standard segmentation pipeline. It can be interpreted as a typical segmentation model of stride 16. This setting is identical to the ``Linear'' baseline in~\cite{strudel2021segmenter}, and we produce similar results by our implementation. Since they have no structural difference with the vanilla ViTs~\cite{dosovitskiy2020image} for image classification, we simply refer to the baseline models as \textbf{ViT-$x$/16} where $x\in$\{Ti, S, M, L\}.

\paragraph{Results} We compare different integrations of vision Transformer for semantic segmentation. Table~\ref{tab:main} summarizes the results. Our RegProxy approach brings consistent performance gains throughout all model sizes with an evidential gap (2\textasciitilde\SI{3}{mIoU} on ADE20K and 2\textasciitilde\SI{4}{mIoU} on Cityscapes), while costing negligible overhead (less than \SI{2}{M} parameters and \SI{1}{GFLOPs} computation for the largest model). It also significantly outperforms the Segmenter~\cite{strudel2021segmenter} which carries a heavy decoder. Moreover, we notice the performance degeneration of Segmenter on smaller models, which is not observed on our approach. As a reference, we also report results from SETR~\cite{zheng2021rethinking} which appends different convolutional decoders to the ViT-L/16 backbone and the UperNet~\cite{xiao2018unified} results ran by us. By these results, we may suggest that our region proxy modeling is a more effective and efficient way to exploit vision Transformers in semantic segmentation compared with the encoder-decoder scheme.

\begin{figure*}
\vspace{-1em}
\centering
\includegraphics[width=1.0\linewidth]{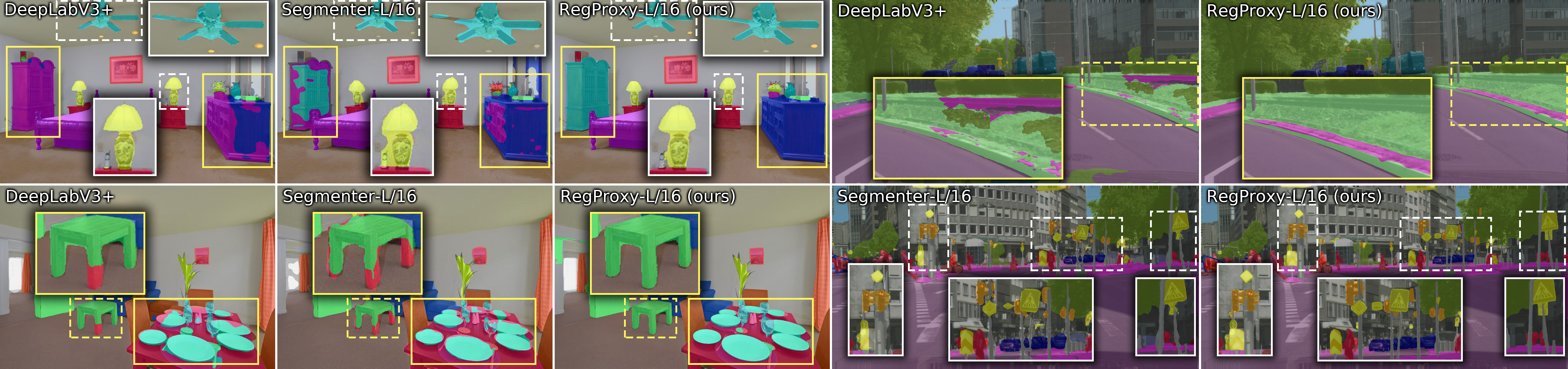}   
\vspace{-1.5em}
\caption{\textbf{Qualitative comparison on ADE20K and Cityscapes.} Our model produces much more accurate contour (in white box) compared to Segmenter~\cite{strudel2021segmenter} and captures more consistent context (in yellow box) compared to DeepLabV3+~\cite{chen2018encoder}. Best viewed zoom in.}
\label{fig:vis}
\vspace{-1em}
\end{figure*}

\subsection{Comparison across Architectures}

\begin{table}[]
\footnotesize
\begin{adjustbox}{max width=1.0\linewidth}
\begin{threeparttable}
\tablestyle{0.9\tabcolsep}{1.2}
\begin{tabular}{l@{\hspace{0.5\tabcolsep}}c|c@{\hspace{0.7\tabcolsep}}c@{\hspace{0.7\tabcolsep}}|c<{\hspace{-\tabcolsep}}>{\hspace{-\tabcolsep}\,}c<{\hspace{-\tabcolsep}}>{\hspace{-\tabcolsep}\,}cc}
method                               & backbone     & FLOPs           & \#params.       & \multicolumn{3}{c}{mIoU}          & FPS            \\ \shline
FCN\cite{long2015fully}              & MobileNetV2  & \phantom{}39.6G & \phantom{0}9.8M & 19.7          & / & -             & 64.4           \\
D.LabV3+\cite{chen2018encoder}       & MobileNetV2  & \phantom{}69.4G & \phantom{}15.4M & 34.0          & / & -             & 43.1           \\ \hline
SegFormer\cite{xie2021segformer}     & MiT-B0       & \phantom{0}8.4G & \phantom{0}3.8M & 37.4          & / & 38.0          & 50.5           \\
Segmenter\cite{strudel2021segmenter} & ViT-Ti/16    & \phantom{0}4.9G & \phantom{0}6.7M & 38.1          & / & 38.8          & -              \\ \hline
\rowcolor{table-bg}
\textbf{RegProxy}                    & ViT-Ti/16    & \phantom{0}3.9G & \phantom{0}5.8M & \textbf{42.1} & / & \textbf{43.1} & 38.9           \\ \hline\hline
% FCN\cite{long2015fully}              & HRNet-W18    & \phantom{0.}19G & \phantom{0.}10M & 36.8          & / & 38.6          & 22.6           \\
OCRNet\cite{yuan2020object}          & HRNet-W18    & \phantom{0.}55G & \phantom{0.}12M & 39.3          & / & 40.8          & 18.9           \\ \hline
SegFormer\cite{xie2021segformer}     & MiT-B1       & \phantom{0.}16G & \phantom{0.}14M & 42.2          & / & 43.1          & 47.7           \\
% UperNet\cite{xiao2018unified,liu2021swin}            & Swin-T       & \phantom{.}236G & \phantom{0.}60M & -             & / & 46.1          & 18.5           \\
Segmenter\cite{strudel2021segmenter} & ViT-S/16     & \phantom{0.}19G & \phantom{0.}26M & 45.3          & / & 46.9          & 29.8           \\ 
MaskFormer\cite{cheng2021per}        & Swin-T       & \phantom{0.}55G & \phantom{0.}42M & 46.7          & / & 48.8          & 22.1           \\ \hline
\rowcolor{table-bg}
\textbf{RegProxy}                    & ViT-S/16     & \phantom{0.}15G & \phantom{0.}22M & \textbf{47.6} & / & 48.5          & 32.1           \\
\rowcolor{table-bg}
\textbf{RegProxy}                    & R26+ViT-S/32 & \phantom{0.}16G & \phantom{0.}36M & \textbf{47.8} & / & \textbf{49.1} & 28.5           \\ \hline\hline
% FCN\cite{long2015fully}              & ResNet-101   & \phantom{.}276G & \phantom{0.}69M & 39.9          & / & 41.4          & 14.8           \\
OCRNet\cite{yuan2020object}          & HRNet-W48    & \phantom{.}165G & \phantom{0.}71M & 43.2          & / & 44.9          & 17.0           \\
D.LabV3+\cite{chen2018encoder}       & ResNet-101   & \phantom{.}255G & \phantom{0.}63M & 45.5          & / & 46.4          & 14.1           \\
% D.LabV3+\cite{chen2018encoder}       & ResNeSt-101  & \phantom{.}263G & \phantom{0.}66M & 46.5          & / & 47.3          & 12.0           \\
D.LabV3+\cite{chen2018encoder}       & ResNeSt-200  & \phantom{.}345G & \phantom{0.}88M & -             & / & 48.4          & -              \\ \hline
% SegFormer\cite{xie2021segformer}     & MiT-B4       & \phantom{0.}96G & \phantom{0.}64M & 50.3          & / & 51.1          & 15.4           \\
Segmenter\cite{strudel2021segmenter} & ViT-B/16     & \phantom{0.}76G & \phantom{.}103M & 48.7          & / & 50.1          & 14.6           \\ \hline
\rowcolor{table-bg}
\textbf{RegProxy}                    & ViT-B/16     & \phantom{0.}59G & \phantom{0.}87M & \textbf{49.8} & / & \textbf{50.5} & 20.1           \\ \hline\hline
DPT\cite{ranftl2021vision}           & DPT-Hybrid   & -               & \phantom{.}123M & -             &   & 49.0          & -              \\
SETR\cite{zheng2021rethinking}       & ViT-L/16     & \phantom{.}422G & \phantom{.}318M & 48.6          & / & 50.1          & \phantom{0}4.5 \\
SegFormer\cite{xie2021segformer}     & MiT-B5       & \phantom{.}183G & \phantom{0.}85M & 51.0          & / & 51.8          & \phantom{0}9.8 \\
Segmenter\cite{strudel2021segmenter} & ViT-L/16     & \phantom{.}370G & \phantom{.}333M & 51.8          & / & \textbf{53.6} & -              \\ \hline
% UperNet\cite{liu2021swin}            & Swin-L       & \phantom{.}647G & \phantom{.}234M & -             & / & 53.5          & \phantom{0}6.2 \\ \hline
\rowcolor{table-bg}
\textbf{RegProxy}                    & R50+ViT-L/32 & \phantom{0.}82G & \phantom{.}323M & 51.0          & / & 51.7          & 12.7           \\
\rowcolor{table-bg}
\textbf{RegProxy}                    & ViT-L/16     & \phantom{.}326G & \phantom{.}306M & \textbf{52.9} & / & 53.4          & \phantom{0}6.6
\end{tabular}\begin{tablenotes}
\footnotesize
    \item[*] All models in the last group except DPT use a larger $640\times 640$ crop.
\end{tablenotes}
\end{threeparttable}
\end{adjustbox}
\vspace{-0.5em}
\caption{\textbf{Comparison to state-of-the-art methods on ADE20K \texttt{val} split.}\label{tab:ade20k} We report both single/multi-scale results. We group the methods based on model capacities and computational costs for fine-grained comparison. Cross group comparison is welcomed.}
\vspace{-1.7em}
\end{table}

\begin{table}[t]
\footnotesize
\centering
\begin{adjustbox}{max width=1.0\linewidth}
\begin{threeparttable}
\tablestyle{0.6\tabcolsep}{1.2}
\begin{tabular}{lc|c@{\hspace{0.7\tabcolsep}}c@{\hspace{0.7\tabcolsep}}c|c<{\hspace{-\tabcolsep}}>{\hspace{-\tabcolsep}\,}c<{\hspace{-\tabcolsep}}>{\hspace{-\tabcolsep}\,}c}
method                               & backbone      & crop     & FLOPs           & \#params.       & \multicolumn{3}{c}{mIoU}          \\ \shline
% FCN\cite{long2015fully}              & MobileNetV2   & full     & \phantom{0}315G & \phantom{0}9.7M & 61.5          & / & -             \\
% D.LabV3+\cite{chen2018encoder}       & MobileNetV2   & full     & \phantom{0}547G & 15.2M           & 75.2          & / & -             \\
D.LabV3+\cite{chen2018encoder}       & ResNet-18      & $769^2$  & \phantom{0}992G & \phantom{0.}12M & 76.3          & / & 77.9          \\
% FCN\cite{long2015fully}              & HRNet-W18S    & full     & \phantom{00}77G & \phantom{0}3.9M & 75.3          & / & -             \\
OCRNet\cite{yuan2020object}          & HRNet-W18     & full     & \phantom{0}424G & \phantom{0.}12M & 78.6          & / & 80.5          \\ \hline
SegFormer\cite{xie2021segformer}     & MiT-B0         & $768^2$  & \phantom{00}52G & \phantom{00.}4M & 75.3          & / & -             \\
SegFormer\cite{xie2021segformer}     & MiT-B1        & $1024^2$ & \phantom{0}244G & \phantom{0.}14M & 78.5          & / & 80.0          \\ \hline
\rowcolor{table-bg}
\textbf{RegProxy}                    & ViT-Ti/16      & $768^2$  & \phantom{00}69G & \phantom{00.}6M & 76.5          & / & 77.7          \\
\rowcolor{table-bg}
\textbf{RegProxy}                    & ViT-S/16      & $768^2$  & \phantom{0}270G & \phantom{0.}23M & \textbf{79.8} & / & \textbf{81.5} \\ \hline\hline
% CCNet\cite{huang2019ccnet}           & ResNet-101    & $769^2$  & 5104G           & \phantom{0.}69M & 79.5          & / & 80.7          \\
OCRNet\cite{yuan2020object}          & HRNet-W48     & full     & 1297G           & \phantom{0.}70M & 80.7          & / & 81.9          \\
Auto-D.Lab\cite{liu2019auto}         & NAS-F48       & $769^2$  & -               & \phantom{0.}44M & -             & / & 80.4          \\
Axial-D.Lab\cite{wang2020axial}      & Axial-D.Lab-XL & -        & 2447G           & \phantom{.}173M & -             & / & 81.1          \\
D.LabV3+\cite{chen2018encoder}       & ResNeSt-200   & full     & -               & \phantom{0.}88M & -             & / & \textbf{82.7} \\ \hline
SETR\cite{zheng2021rethinking}       & ViT-B/16      & $768^2$  & -               & \phantom{0.}98M & 79.5          & / & -             \\
SETR\cite{zheng2021rethinking}       & ViT-L/16      & $768^2$  & -               & \phantom{.}318M & 79.3          & / & 82.2          \\
Segmenter\cite{strudel2021segmenter} & ViT-B/16      & $768^2$  & 1344G           & \phantom{.}103M & -             & / & 80.6          \\
Segmenter\cite{strudel2021segmenter} & ViT-L/16      & $768^2$  & -               & \phantom{.}337M & 79.1          & / & 81.3          \\ \hline
\rowcolor{table-bg}
\textbf{RegProxy}                    & ViT-B/16      & $768^2$  & 1064G           & \phantom{0.}88M & 81.0          & / & 82.2          \\
\rowcolor{table-bg}
\textbf{RegProxy}                    & ViT-L/16      & $768^2$  & -               & \phantom{.}307M & \textbf{81.4} & / & \textbf{82.7}
\end{tabular}
\end{threeparttable}
\end{adjustbox}
\vspace{-0.5em}
\caption{\textbf{Comparison to state-of-the-art methods on Cityscapes \texttt{val} split.}\label{tab:cityscapes} The ``full'' crop indicates the whole image inference, while others indicate the \textit{sliding window} protocol.}
\vspace{-1.7em}
\end{table}

\paragraph{ADE20K} We conduct comprehensive comparisons with the state-of-the-art methods on ADE20K. Results in Figure~\ref{fig:perf-gflops-param} show that the RegProxy is among the most competitive models in terms of performance-efficiency trade-off. RegProxy consistently outperforms the state-of-the-art Segmenter~\cite{strudel2021segmenter} by a large margin, which uses the same ViT backbones and pre-training as ours. It also shows significant superiority compared with recent SegFormer~\cite{strudel2021segmenter}, SwinTransformer~\cite{liu2021swin} and SETR~\cite{zheng2021rethinking}. Table~\ref{tab:ade20k} gives a more detailed comparison with respect to parameters, GFLOPs, inference speed and performance. The smallest RegProxy-Ti/16 achieves {\SI{42.1}{mIoU}} with only {\SI{5.8}{M}} parameters and {\SI{3.9}{GFLOPs}}, and runs at a speed of {\SI{38.9}{FPS}}, which outperforms the best efficient models by large margins (greater than \SI{4.0}{mIoU}). Our small model RegProxy-S/16 has surpassed the heaviest CNN model (DeepLabV3+ w/ ResNeSt-200) using only {25\%} parameters and {4\%} computation. The medium sized RegProxy-B/16 significantly outperforms all CNN models with much less computation. The largest RegProxy-L/16 achieves {\SI{52.9}{mIoU}} \emph{without} multi-scale inference, outperforms Segmenter~\cite{strudel2021segmenter} by \SI{1.1}{mIoU}. The hybrid models also achieve notable results even with a large-stride backbone. The RegProxy-[R26+S/32] achieves 47.8/\SI{49.1}{mIoU}, and the RegProxy-[R50+L/32] achieves a comparable 51.0/\SI{51.7}{mIoU} among jumbo Transformer-based models using only 1/4 computations. And generously, our RegProxy runs at a significantly higher frame rate among different sized models thanks to the concise design.

\paragraph{Cityscapes and Pascal Context} In Table~\ref{tab:cityscapes} we compare the state-of-the-art methods on Cityscapes. We observe similar results as on ADE20K. Our RegProxy-\{Ti/16, S/16\} outperform their counterparts with a \textasciitilde\SI{1.2}{mIoU} margin. Our RegProxy-\{B/16, L/16\} also achieve state-of-the-art performances compared to larger models. Notably, they outperform SETR~\cite{zheng2021rethinking} and Segmenter~\cite{strudel2021segmenter} by \textasciitilde\SI{2}{mIoU} using identical backbones. We notice that on Cityscapes, the performance gained from model capacities tends to saturate on larger models. RegProxy-L/16 is only \SI{0.4}{mIoU} higher than RegProxy-B/16 (the gap on ADE20K is \SI{2.2}{mIoU} using both $512\times 512$ crops). This is also observed on SETR~\cite{zheng2021rethinking} and Segmenter~\cite{strudel2021segmenter}. We hypothesize that this is due to the simpler context of Cityscapes. We report results on Pascal Context in Table~\ref{tab:pascal_context}. RegProxy-L/16 achieves \SI{58.4}{mIoU} and significantly outperforms the state-of-the-art CNN models, and achieves comparable performance with Segmenter~\cite{strudel2021segmenter} using less resources. We present qualitative comparison in Figure~\ref{fig:vis} and appendix.

\begin{table}[]
\vspace{-0.5em}
\footnotesize
\centering
\begin{adjustbox}{max width=1.0\linewidth}
\begin{threeparttable}
\tablestyle{1.0\tabcolsep}{1.15}
\begin{tabular}{lc|cc|c<{\hspace{-\tabcolsep}}>{\hspace{-\tabcolsep}\,}c<{\hspace{-\tabcolsep}}>{\hspace{-\tabcolsep}\,}c}
method                               & backbone   & FLOPs          & \#params.      & \multicolumn{3}{c}{mIoU}          \\ \shline
% FCN\cite{long2015fully}              & ResNet-101 & 242G           & \phantom{0}69M & 44.1          & / & 45.3          \\
DeepLabV3+\cite{chen2018encoder}     & ResNet-101 & 224G           & \phantom{0}63M & 47.2          & / & 48.3          \\
OCRNet\cite{yuan2020object}          & HRNet-W48  & 144G           & \phantom{0}71M & -             & / & 56.2          \\ \hline
SETR\cite{zheng2021rethinking}       & ViT-L/16   & 238G           & 318M           & 54.9          & / & 55.8          \\
Segmenter\cite{strudel2021segmenter} & ViT-B/16   & \phantom{0}67G & 102M           & -             & / & 55.0          \\
Segmenter\cite{strudel2021segmenter} & ViT-L/16   & 210G           & 333M           & 58.1          & / & \textbf{59.0} \\ \hline
\rowcolor{table-bg}
\textbf{RegProxy}                    & ViT-B/16   & \phantom{0}52G & \phantom{0}87M & 55.2          & / & 55.4          \\
\rowcolor{table-bg}
\textbf{RegProxy}                    & ViT-L/16   & 183G           & 305M           & \textbf{58.4} & / & 58.8         
\end{tabular}
\end{threeparttable}
\end{adjustbox}
\vspace{-0.5em}
\caption{\textbf{Comparison to state-of-the-art methods on Pascal Context \texttt{val} split.}\label{tab:pascal_context}}
\vspace{-1.7em}
\end{table}

\subsection{Analysis and Ablation Study}
\label{sec:analysis}

In this section, we present the most significant analysis and ablation studies. Due to limited space, we present more of them in the appendix.

\begin{figure}[b]
\vspace{-1.5em}
\centering
\begin{subfigure}[t]{0.45\linewidth}
    \centering
    \includegraphics[width=1.0\linewidth]{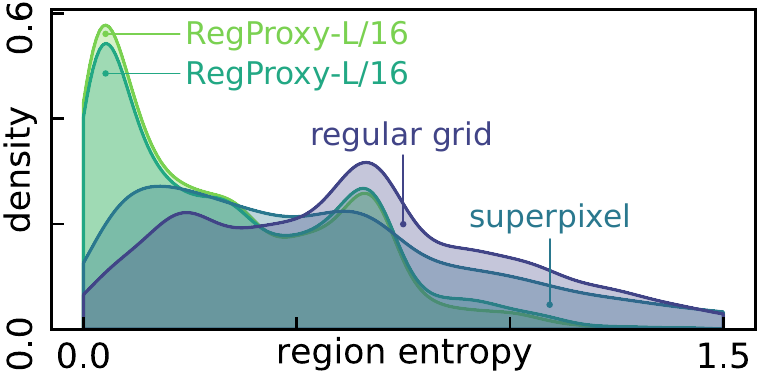}
    % \caption{}
    % \label{fig:entropy_hist}
\end{subfigure}
\hspace{0.01\linewidth}
\begin{subfigure}[t]{0.47\linewidth}
    \centering
    \includegraphics[width=1.0\linewidth]{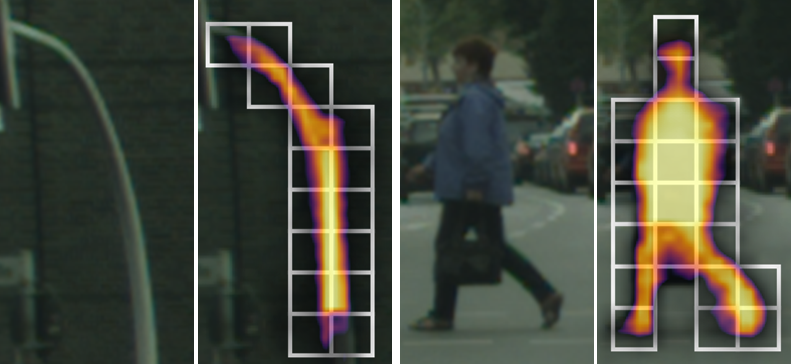}
    % \caption{}
\end{subfigure}
\vspace{-0.5em}
\caption{\textbf{(Left) Distribution of region entropies}\label{fig:entropy_hist} estimated by kernel density estimation (KDE). Regions with entropy of 0 are ignored.
\textbf{(Right) Geometrics of the \emph{class-agnostic} regions} and its corresponding tokens (marked using white cell).}
\vspace{-0.5em}
\end{figure}

\paragraph{Region Semantics}
We analyze the semantical homogeneity of regions, which is described using ``region entropy'': we calculate the per-pixel category histogram within every regions using Eq.~\ref{eq:class-hist} and compute their entropies~\cite{shannon1948mathematical}. We analyze the distribution of region entropies on the entire ADE20K \texttt{val} split, with regard to three sources: regular cell used by conventional segmentation model such as~\cite{strudel2021segmenter}, superpixels used in our pilot study, and learned regions from our RegProxy approach. The results are shown in Figure~\ref{fig:entropy_hist}. Although we do not apply any explicit regularization, the learned regions still exhibit highly compact semantics compared to regular cells and superpixels. 
% We also find that larger models learns more uniformed regions.
In the right part of Figure~\ref{fig:entropy_hist}, we visualize the leaned probabilistic regions of a number of tokens on tiny crops of Cityscapes validation images. Note the regions are \emph{class-agnostic} (as association of pixels and nearby tokens) and leaned from shallow features. However, they still capture fine-grained boundaries of high level classes. This suggest that the computation primitives in our model (i.e., tokens) carries more uniformed semantics compared to CNNs or other vision Transformer models that compute on structured features, which probably leads to easier optimization hence the better performances.

\begin{figure}[t]
\vspace{-1.5em}
\centering
\includegraphics[width=1.0\linewidth]{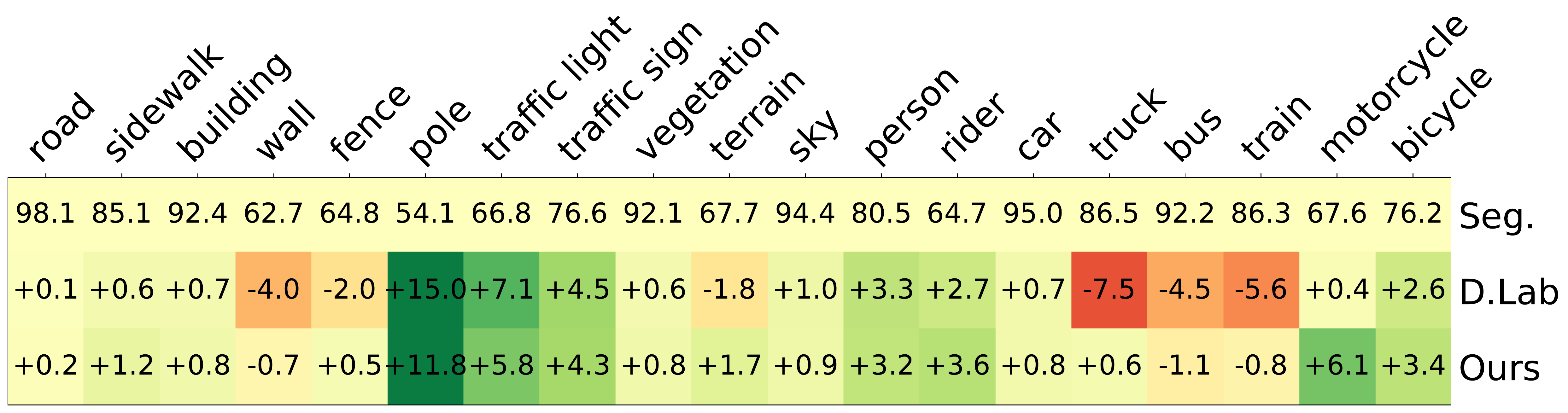}
\vspace{-2.0em}
\caption{\textbf{Per-class performance on Cityscapes.}\label{fig:city-heat} We report relative mIoU of DeepLabV3+ and our \textbf{RegProxy-L/16} model compared with Segmenter-L/16~\cite{strudel2021segmenter}.}
\vspace{-1.5em}
\end{figure}

\paragraph{Per-Class Performance}
In Figure~\ref{fig:city-heat}, we analyze per-class performance on Cityscapes. Compared to Transformer-based Segmenter~\cite{strudel2021segmenter}, DeepLabV3+~\cite{chen2018encoder} is good at handling small/thin classes (e.g., traffic sign, pole) thanks to its small stride, yet fails more on confusing classes (e.g., bus, wall) which require wider context. Our region proxy modeling possesses both of their merits and consistently outperforms or being comparable to both of the methods on all classes. This is credited to the \emph{early region} design that captures fine regions in advance and models region-wise relations afterwards, hence learns global context without loss of details.

\paragraph{Depth of Token Head} As illustrated in Figure~\ref{fig:stem-affinity-head}, we use the first $M$ Transformer layers in the token head. $M$ is set to 0 for our hybrid models since the convolutional stem is strong enough for region learning and embedding. Here we investigate the $M$ setting for vanilla ViT based models. The results are displayed in Table~\ref{tab:abliation-feature-stage}. We find that using early layers (e.g., 3\textasciitilde5 layers) generously works well, while setting $M$ to 0 (a too shallow token head) will harm the performance. Interestingly and importantly, we also find that a too deep token head will cause a significant performance drop. To the extreme, setting $M=12$ (uses all Transformer layers, the proxy head degenerates to a typical decoder) will yield a performance no better than the baseline's. This indicates the importance of the Transformer layers \emph{after} the proxy head, which model context for learned regions.

\begin{table}[h]
\vspace{-0.6em}
\centering
\footnotesize
\begin{adjustbox}{max width=1.0\linewidth}
\tablestyle{0.6\tabcolsep}{1.15}
\begin{tabular}{l@{\hspace{0.8\tabcolsep}}|ccccccc|c}
depth $M$  & 0                           & 3             & 4    & 5                           & 6                           & 9                           & 12                          & baseline \\ \shline
ADE20K     & {\color[HTML]{9B9B9B} 46.3} & 47.1          & 47.0 & \textbf{47.2}               & {\color[HTML]{9B9B9B} 46.8} & {\color[HTML]{9B9B9B} 46.6} & {\color[HTML]{9B9B9B} 45.7} & 45.0     \\
Cityscapes & {\color[HTML]{9B9B9B} 76.8} & \textbf{79.0} & 78.8 & {\color[HTML]{9B9B9B} 78.2} & {\color[HTML]{9B9B9B} 77.7} & {\color[HTML]{9B9B9B} 76.1} & {\color[HTML]{9B9B9B} 75.2} & 75.4    
\end{tabular}
\end{adjustbox}
\vspace{-1em}
\caption{\textbf{Ablation on depth of token features used for region learning}. We report single scale mIoU results of RegProxy-S/16 on ADE20K and Cityscapes using half of the training schedule.}
\label{tab:abliation-feature-stage}
\vspace{-1.7em}
\end{table}

\vspace{-0.1em}
\section{Conclusion}
\vspace{-0.1em}

In this paper, we present \emph{region proxy}, a novel and efficient modeling of semantic segmentation. It interprets the image as a tessellation of learnable regions, each of which has flexible geometrics and carries homogeneous semantics. We conduct semantic segmentation by per-region prediction on top of region embeddings, which are encoded using Transformer in a sequence-to-sequence fashion. Without a decoder, the RegProxy segmentation models still exhibit the most competitive performance-efficiently trade-off among its dense prediction counterparts. We hope our region proxy modeling provide an inspiring perspective of efficient image representation for semantic segmentation and other vision tasks.

\vspace{-0.1em}
\section{Acknowlegement}
\vspace{-0.1em}

This work was supported by the National Key Research and Development Project of China (No. 2021ZD0110700), Shanghai Municipal Science and Technology Major Project (2021SHZDZX0102), Shanghai Qi Zhi Institute, and SHEITC (2018-RGZN-02046).

%%%%%%%%% REFERENCES
\pagebreak
{\small
\bibliographystyle{ieee_fullname}
\bibliography{egbib}
}

\clearpage
\appendix

\section{Preliminaries of Transformer}

% \subsection{Multi-Head Self-Attention}

We provide details of the Multi-Head Attention (MSA) block we mentioned in Section~\ref{sec:transformer-region-encoder} for exhaustivity. ViT~\cite{dosovitskiy2020image} inherits the exact $\mathbf{qkv}$ self-attention proposed in\cite{vaswani2017attention}. The query, key and value $\mathbf{q}, \mathbf{k}, \mathbf{v}\in\mathbf{R}^{N\times D_h}$ are linearly projected from the input tokens $\mathbf{X}\in\mathbb{R}^{N\times D}$. Each output token is a weighted sum over all values $\mathbf{v}$ in the sequence, where the weights $A_{ij}$ are based on the pairwise similarity between two elements in the sequence with respect to their $\mathbf{q}_i,\mathbf{k}_j$ representations. Self-attention can be formulated as:
{\small
\begin{align}
  \begin{bmatrix} \mathbf{q}, \mathbf{k}, \mathbf{v} \end{bmatrix} &= \mathbf{X}\mathbf{U}_\textit{qkv} &  \mathbf{U}_\textit{qkv} &\in\mathbb{R}^{D\times3D_h}, \\
  A &= \operatorname{softmax}\left(\mathbf{q}\mathbf{k}^\top/\sqrt{D_h}\right) & A &\in \mathbb{R}^{N\times N}, \\
  \operatorname{SA}(\mathbf{X}) &= A\mathbf{v}.
\end{align}}
Multi-head self-attention (MSA) is an extension of the self-attention where $k$ $\operatorname{SA}$ heads are applied to the input sequence in parallel. It returns a linear projection of the concatenated outputs of the $\operatorname{SA}$s:
{\small
\begin{equation}
  \operatorname{MSA}(\mathbf{X}) = \begin{bmatrix}
    \operatorname{SA}_1(\mathbf{X}), \operatorname{SA}_2(\mathbf{X}), \dots, \operatorname{SA}_k(\mathbf{X})
  \end{bmatrix}\mathbf{U}_\textit{msa},
\end{equation}}
where $\mathbf{U}_\textit{msa}\in\mathbb{R}^{(kD_h)\times D}$.

\section{More Implementation Details}

\begin{algorithm}
	\caption{Code of ``painting'' in a PyTorch-like style.}
	\label{alg:code}
	\algcomment{\fontsize{7.2pt}{0em}\selectfont \texttt{rar}: rearrange of dimensions; \texttt{mm}: matrix multiplication.
		%\vspace{-1.em}
	}
	\definecolor{codeblue}{rgb}{0.25,0.5,0.5}
	\lstset{
		backgroundcolor=\color{white},
		basicstyle=\fontsize{7.2pt}{7.2pt}\ttfamily\selectfont,
		% columns=fullflexible,
		breaklines=true,
		captionpos=b,
		commentstyle=\fontsize{7.2pt}{7.2pt}\color{codeblue},
		keywordstyle=\fontsize{7.2pt}{7.2pt},
    % showtabs=true,
    % showspaces=false,
    showstringspaces=false
  }
	\def\redcolor{\color{red}}
	\def\blackcolor{\color{black}}
\begin{lstlisting}[language=python,escapechar=@]
# y: (B, N, K)       # The token logits.
# Q: (B, nhw, H, W)  # The affinity map Q. n = 9.

# get neighbors for each cell
y = rar(y, "B N K -> B K H W")
nb = im2col(y, kernel_size=3, padding=1)
nb = rar(nb, "B (K n) (H W) -> B H W n K")

# produce output logits map
Q = rar(Q, "B (n h w) H W -> B H W (h w) n)
out = mm(Q, nb)
out = rar(out, "B H W (h w) K -> B (Hh) (Ww) K")
\end{lstlisting}
\end{algorithm}

% \subsection{Painting}

In Algorithm~\ref{alg:code}, We provide the psuedo code of the ``painting'' process in Section~\ref{sec:per-retion-prediction}, i.e., producing the final logits map from the token logits and the predicted pixel-token affinity map $\mathbf{Q}$.

\section{Extended Ablation Study}

In this section, we provide more ablation results as a supplementary of Section~\ref{sec:analysis} in the main paper. All the experiments are conducted using half of the training schedule (i.e., 80k for ADE20K and 40k for Cityscapes) unless otherwise specified. We report median value of three runs.

\paragraph{Output Strides}
Thanks to the class-agnostic region design, we can attain output segmentation of \emph{arbitrary resolutions} with negligible overheads. Here we investigate the effect of output stride, which is determined by the $h,w$ mentioned in Section~\ref{sec:pixel-token-association}. We report results in Table~\ref{tab:out-strides}.

\begin{table}[h]
\vspace{-0.5em}
\footnotesize
\centering
\begin{adjustbox}{max width=1.0\linewidth}
\begin{threeparttable}
\tablestyle{0.4\tabcolsep}{1.15}
\begin{tabular}{l|cc|c@{\hspace{0.5\tabcolsep}}c@{\hspace{0.5\tabcolsep}}|cc}
arch.                                                                          & $h\times w$              & stride                   & GFLOPs                                 & \#params.                              & ADE20K                       & City.                        \\ \shline
{\color[HTML]{9B9B9B} ViT-Ti/16}                                               & {\color[HTML]{9B9B9B} -} & {\color[HTML]{9B9B9B} -} & {\color[HTML]{9B9B9B} \phantom{0}3.8G} & {\color[HTML]{9B9B9B} \phantom{0}5.7M} & {\color[HTML]{9B9B9B} 38.76} & {\color[HTML]{9B9B9B} 72.02} \\ \hline
                                                                               & $1\times1$               & $\times16$               & \phantom{0}3.9G                        & \phantom{0}5.8M                        & 39.30                        & 73.11                        \\
                                                                               & $2\times2$               & $\times8$\phantom{$1$}   & +0.0G                                  & +0.0M                                  & 40.23                        & 74.96                        \\
                                                                               & $4\times4$               & $\times4$\phantom{$1$}   & +0.0G                                  & +0.0M                                  & \textbf{40.76}               & 75.08                        \\
                                                                               & $8\times8$               & $\times2$\phantom{$1$}   & +0.1G                                  & +0.1M                                  & 40.51                        & \textbf{75.31}               \\
\multirow{-5}{*}{RegProxy-Ti/16}                                               & $16\times16$             & $\times1$\phantom{$1$}   & +0.4G                                  & +0.4M                                  & \textbf{41.01}               & \textbf{75.15}               \\ \hline\hline
{\color[HTML]{9B9B9B} ViT-S/16}                                                & {\color[HTML]{9B9B9B} -} & {\color[HTML]{9B9B9B} -} & {\color[HTML]{9B9B9B} 14.9G}           & {\color[HTML]{9B9B9B} 22.0M}           & {\color[HTML]{9B9B9B} 45.04} & {\color[HTML]{9B9B9B} 75.39} \\ \hline
                                                                               & $1\times1$               & $\times16$               & 15.1G                                  & 22.2M                                  & 45.81                        & 76.71                        \\
                                                                               & $2\times2$               & $\times8$\phantom{$1$}   & +0.0G                                  & +0.0M                                  & 46.91                        & 78.48                        \\
                                                                               & $4\times4$               & $\times4$\phantom{$1$}   & +0.1G                                  & +0.1M                                  & 47.18                        & \textbf{78.71}               \\
                                                                               & $8\times8$               & $\times2$\phantom{$1$}   & +0.2G                                  & +0.2M                                  & \textbf{47.20}               & 78.37                        \\
\multirow{-5}{*}{RegProxy-S/16}                                                & $16\times16$             & $\times1$\phantom{$1$}   & +0.8G                                  & +0.9M                                  & \textbf{47.34}               & \textbf{78.81}               \\ \hline\hline
                                                                               & $1\times1$               & $\times32$               & 16.2G                                  & 36.3M                                  & 45.40                        & 74.66                        \\
                                                                               & $2\times2$               & $\times16$               & +0.0G                                  & +0.0M                                  & 46.97                        & 77.56                        \\
                                                                               & $4\times4$               & $\times8$\phantom{$1$}   & +0.1G                                  & +0.1M                                  & \textbf{48.11}               & \textbf{78.18}               \\
                                                                               & $8\times8$               & $\times4$\phantom{$1$}   & +0.2G                                  & +0.2M                                  & \textbf{47.79}               & 78.05                        \\
\multirow{-5}{*}{\begin{tabular}[c]{@{}l@{}}RegProxy-\\ R26+S/32\end{tabular}} & $16\times16$             & $\times2$\phantom{$1$}   & +0.3G                                  & +0.9M                                  & 47.42                        & \textbf{78.34}              
\end{tabular}
\end{threeparttable}
\end{adjustbox}
\vspace{-0.5em}
\caption{\textbf{Comparison of different output strides.}\label{tab:out-strides} We report single scale results on ADE20K and Cityscapes. The GFLOPs are evaluated on $512\times 512$ crops. In {\color[HTML]{9B9B9B} gray} are the linear baselines. We \textbf{bold} the top-2 entries for each model.}
\vspace{-0.5em}
\end{table}

Despite the almost free cost of attaining high resolution results, it is not always the best choice to set a large $(h, w)$. Part of the reason is that the performance upper bound tends to saturate as the prediction gets finer. We also hypothesize that a too large $(h, w)$ makes the model harder to train. In the main paper, we report results with $(h,w)$ set to $(4,4)$ to align the output stride with common segmentation models~\cite{chen2018encoder, yuan2020object}. Still, increasing $(h,w)$ is a good choice which generally improves the performance with negligible cost.

\paragraph{The $3\times3$ Conv} The $3\times3$ depth-wise convolution in the affinity head was initially introduced to fuse local information for region geometrics prediction. We investigate its effect in Table~\ref{tab:3x3conv}. We find it improves the performance on ADE20K, while has no significant effect on Cityscapes. We also notice a normal convolution (with group of $1$) yields similar results with the depth-wise one. To sum up, the $3\times3$ depth-wise convolution in the affinity head improves the performance, however it is not a determinative component. Use early transformer layers alone can also achieve considerable performances. This is reasonable since it is only used for region geometrics prediction, while is not involved in the actual context modeling.

\begin{table}[h]
\vspace{-0.5em}
\footnotesize
\centering
\begin{adjustbox}{max width=1.0\linewidth}
\begin{threeparttable}
\tablestyle{0.5\tabcolsep}{1.15}
\begin{tabular}{l|cc|c@{\hspace{0.5\tabcolsep}}c@{\hspace{0.5\tabcolsep}}|cc}
arch.                           & $3\times3$ conv. & dpt.\tnote{$\dagger$}       & GFLOPs          & \#params.       & ADE20K         & City.          \\ \shline
\multirow{3}{*}{RegProxy-Ti/16} &                  &            & \phantom{0}3.9G & \phantom{0}5.7M & 40.01          & 75.04          \\
                                & \checkmark       &            & +0.3G           & +0.4M           & 40.72          & 74.96          \\
                                & \checkmark       & \checkmark & +0.0G           & +0.1M           & \textbf{40.76} & \textbf{75.08} \\ \hline
\multirow{3}{*}{RegProxy-S/16}  &                  &            & 15.0G           & 22.1M           & 46.78          & 78.63          \\
                                & \checkmark       &            & +1.3G           & +1.3M           & 46.98          & 78.68          \\
                                & \checkmark       & \checkmark & +0.1G           & +0.2M           & \textbf{47.18} & \textbf{78.71}
\end{tabular}
\begin{tablenotes}
  \item[$\dagger$] Whether to use depth-wise convolution~\cite{chollet2017xception}.
\end{tablenotes}
\end{threeparttable}
\end{adjustbox}
\vspace{-0.5em}
\caption{\textbf{Effect of the $3\times3$ conv.}\label{tab:3x3conv}}
\vspace{-0.5em}
\end{table}

\paragraph{Pre-training}

We study the effect of different ViT pre-trainings. We initialize our model using three settings: \begin{inparaenum}[1)]
  \item Random initialization;
  \item DeiT~\cite{touvron2021training} pre-training on ImageNet-1k;
  \item AugReg~\cite{steiner2021train} pre-training on ImageNet-21k following recent Segmenter~\cite{strudel2021segmenter}.
\end{inparaenum} 
On RegProxy-S/16, we also report results using DINO~\cite{caron2021emerging} self-supervised pre-training. Table~\ref{tab:pretrain} summarizes the results. Similar to many recent works~\cite{strudel2021segmenter, liu2021swin, zheng2021rethinking}, RegProxy benefits from pre-training on large-scale image dataset, while a random initialization will lead to a dramatic performance drop. However, the RegProxy model without pre-training still performs better than its counterpart reported in~\cite{strudel2021segmenter} (\SI{18.83}{mIoU} \textit{vs.} \SI{4.42}{mIoU} on ADE20K, both using ViT-S/16 backbone).

\begin{table}[h]
\vspace{-0.5em}
\footnotesize
\centering
\begin{adjustbox}{max width=1.0\linewidth}
\begin{threeparttable}
\tablestyle{0.5\tabcolsep}{1.15}
\begin{tabular}{l|lcc|cc}
arch.                                                       & \multicolumn{1}{c}{pre-train}  & IN-21k                   & self-sup.                & ADE20K                                 & Cityscapes               \\ \shline
                                                            & none                           & -                        & -                        & 13.96                                  & 42.58                    \\
                                                            & DeiT\cite{touvron2021training} &                          &                          & 39.42                                  & 74.56                    \\
\multirow{-3}{*}{RegProxy-Ti/16}                            & ViT\cite{steiner2021train}     & \checkmark               &                          & \textbf{40.76}                         & \textbf{75.08}           \\ \hline\hline
{\color[HTML]{9B9B9B} Segmenter\cite{strudel2021segmenter}} & {\color[HTML]{9B9B9B} none}    & {\color[HTML]{9B9B9B} -} & {\color[HTML]{9B9B9B} -} & {\color[HTML]{9B9B9B} \phantom{0}4.42} & {\color[HTML]{9B9B9B} -} \\ \hline
                                                            & none                           & -                        & -                        & 18.83                                  & 49.36                    \\
                                                            & DeiT\cite{touvron2021training} &                          &                          & 45.73                                  & 77.69                    \\
                                                            & DINO\cite{caron2021emerging}   &                          & \checkmark               & 42.21                                  & 77.57                    \\
\multirow{-4}{*}{RegProxy-S/16}                             & ViT\cite{steiner2021train}     & \checkmark               &                          & \textbf{47.18}                         & \textbf{78.71}          
\end{tabular}
\end{threeparttable}
\end{adjustbox}
\vspace{-0.5em}
\caption{\textbf{Performances using different pre-training.}\label{tab:pretrain}}
\vspace{-0.5em}
\end{table}

\section{More Experimental Results}

\paragraph{Region Semantics Regularization} We find adding explicit regularization with respect to the semantical homogeneity of the learned regions will harm the performance. The approach is to minimize the $L_2$-norm of the region category histogram\footnote{This is also used in Section~\ref{sec:analysis} to calculate the region entropies.}:
\begin{equation}
  \operatorname{hist}\left(\mathbf{s}\right) = L_1\left(\sum_{\mathbf{p}:\mathbf{s}\in\mathcal{N}_\mathbf{p}}q_\mathbf{s}(\mathbf{p})\cdot\operatorname{onehot}\left(\hat{y}(\mathbf{p})\right)\right),
\end{equation}
where $\hat{y}(\mathbf{p})$ is the ground truth of pixel $\mathbf{p}$. The results are shown in Table~\ref{tab:regularization}. The models with explicit regularization perform worse with evidential gaps.

\begin{table}[h]
\vspace{-0.5em}
\footnotesize
\centering
\begin{adjustbox}{max width=1.0\linewidth}
\begin{threeparttable}
\tablestyle{0.8\tabcolsep}{1.15}
\begin{tabular}{l|c|cc}
arch.                           & w/ regularization & ADE20K         & Cityscapes     \\ \shline
\multirow{2}{*}{RegProxy-Ti/16} &                   & \textbf{40.76} & \textbf{75.08} \\
                                & \checkmark        & 40.66          & 74.45          \\ \hline
\multirow{2}{*}{RegProxy-S/16}  &                   & \textbf{47.18} & \textbf{78.71} \\
                                & \checkmark        & 46.38          & 78.02         
\end{tabular}
\end{threeparttable}
\end{adjustbox}
\vspace{-0.5em}
\caption{\textbf{Effect of explicit regularization on region semantics.}\label{tab:regularization}}
\vspace{-0.5em}
\end{table}

\paragraph{Multi-Level Features} As a common technique, using multi-level feature for token logits prediction also improves the performance of our RegProxy models. Specifically, we feed the concatenated tokens features (of layer $L/2$, $3L/4$ and $L$) to the linear classifier, instead of using the output tokens of the last layer. The results are reported in Table~\ref{tab:multi-feat}. However, the improvements are marginal, hence in the main paper, we report results without features concatenation.

\begin{table}[h]
\vspace{-0.5em}
\footnotesize
\centering
\begin{adjustbox}{max width=1.0\linewidth}
\begin{threeparttable}
\tablestyle{0.8\tabcolsep}{1.15}
\begin{tabular}{l|c|cc}
arch.                           & multi-level feat. & ADE20K         & Cityscapes     \\ \shline
\multirow{2}{*}{RegProxy-Ti/16} &                   & 40.76          & 75.08          \\
                                & \checkmark        & \textbf{40.98} & \textbf{75.54} \\ \hline
\multirow{2}{*}{RegProxy-S/16}  &                   & 47.18          & 78.71          \\
                                & \checkmark        & \textbf{47.31} & \textbf{78.88}
\end{tabular}
\end{threeparttable}
\end{adjustbox}
\vspace{-0.5em}
\caption{\textbf{Effect of predicting on concatenated token features.}\label{tab:multi-feat}}
\vspace{-1em}
\end{table}

% \paragraph{Other Backbones}

\section{More Qualitative Results}

\paragraph{Geometrics of the Leaned Regions}

\begin{figure*}[t]
  \centering
  \includegraphics[width=1.0\linewidth]{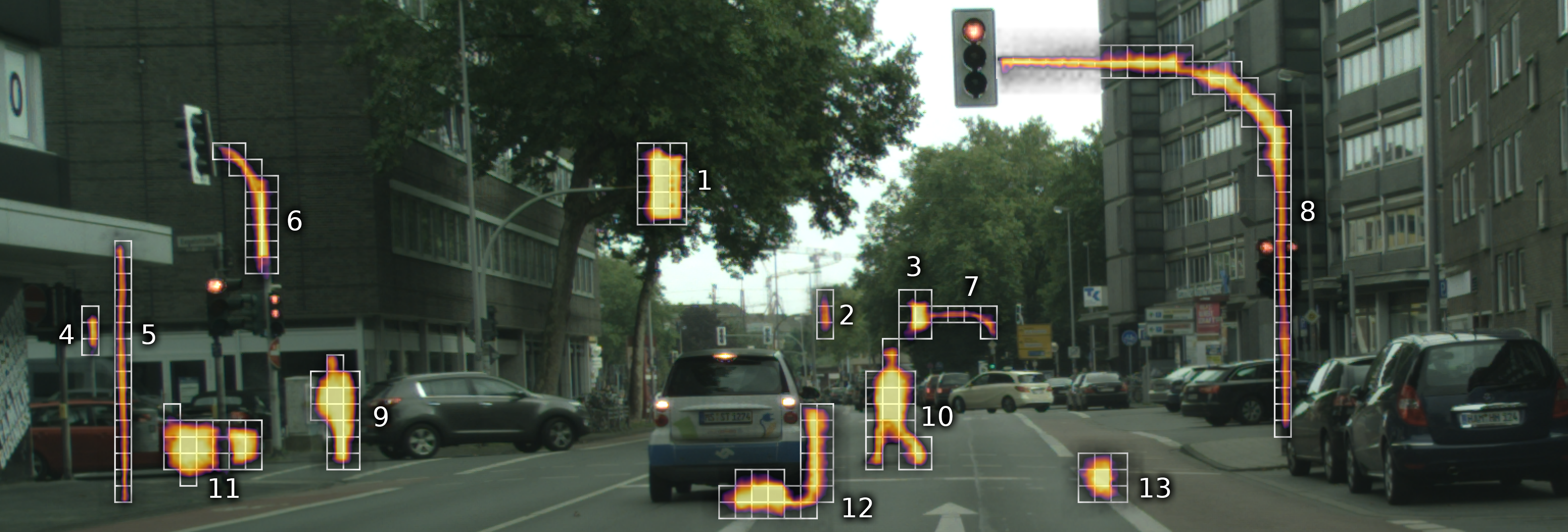}
  \vspace{-1.5em}
  \caption{\textbf{Geometrics of the leaned \emph{class-agnostic regions}} and its corresponding tokens (marked using white cell) on a Cityscapes validation image.\label{fig:leaned-regions} We identify the \emph{token class} for better interpretation: 1\textasciitilde3: \texttt{traffic light}; 4: \texttt{traffic sign}; 5\textasciitilde8: \texttt{pole}; 9\textasciitilde10: \texttt{person}; 11: part of \texttt{car}; 12: surrounding \texttt{road} tokens of a car; 13: inner regions of \texttt{road}.}
  \vspace{-0.5em}
\end{figure*}%

As a supplementary of Section~\ref{sec:analysis}, in Figure~\ref{fig:leaned-regions}, we select a few tokens that have been classified to specific classes, and visualize their corresponding regions. Note the region geometrics is class-agnostic. The heat map is acquired by stacking the probabilistic region descriptions. The learned regions capture fine-grained boundaries, even for small/thin classes such as \texttt{pole} and \texttt{traffic light} and complicated classes such as \texttt{person}. For tokens that locate at the deep inside of the semantics areas (e.g., No.13 in Figure~\ref{fig:leaned-regions}), their corresponding regions are close to Gaussian masks.

\paragraph{Visualization of the Segmentation Results}

\begin{figure*}[]
  \vspace{-1em}
  \centering
  \includegraphics[width=1.0\linewidth]{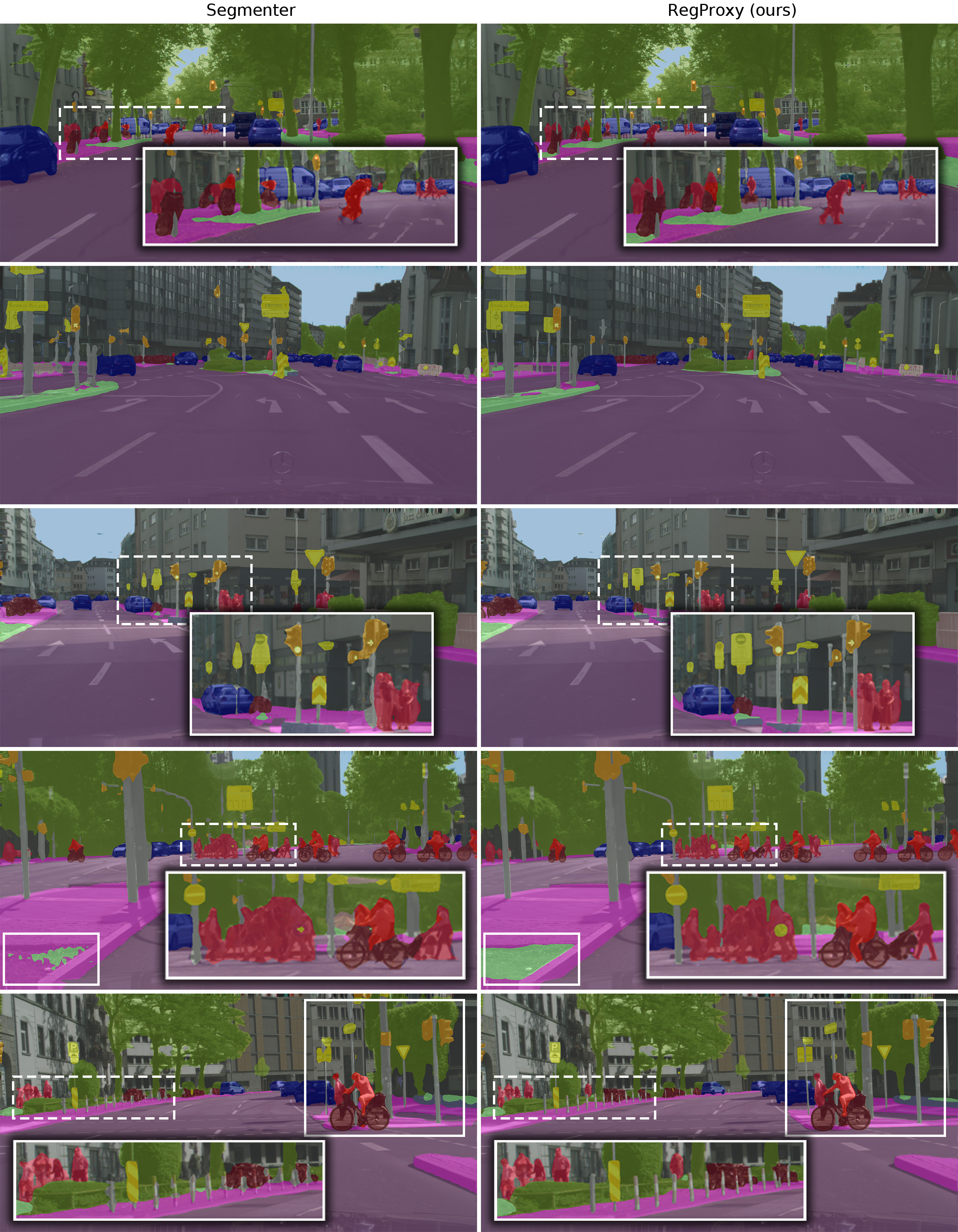}
  \vspace{-1.5em}
  \caption{\textbf{Qualitative comparison on Cityscapes.}\label{fig:more-vis-cityscapes}}
  \vspace{-0.5em}
\end{figure*}

\begin{figure*}[]
  \centering
  \includegraphics[width=1.0\linewidth]{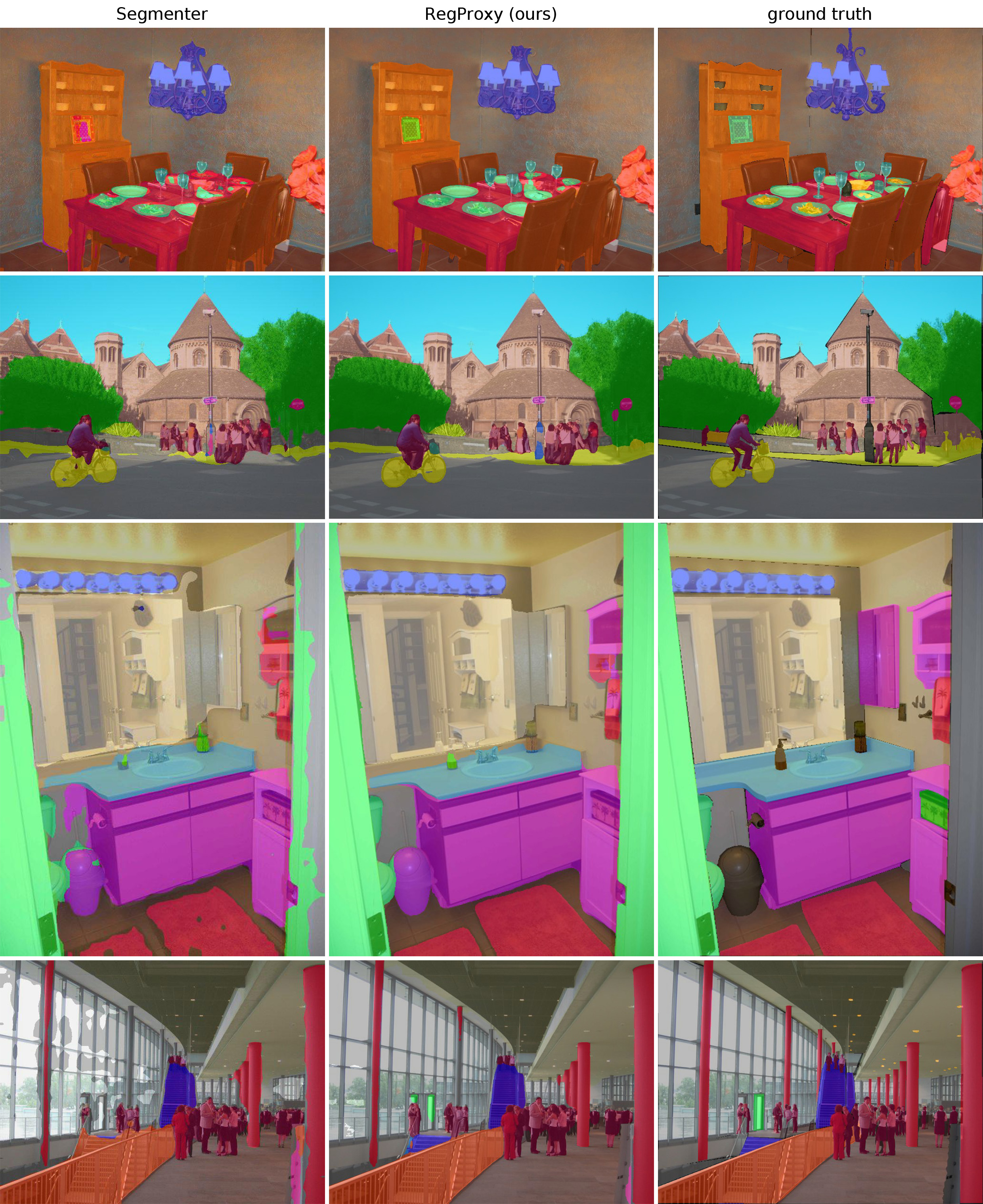}
  \vspace{-1.5em}
  \caption{\textbf{Qualitative comparison on ADE20K.}\label{fig:more-vis-ade20k}}
  \vspace{-0.5em}
\end{figure*}

We provide more qualitative comparisons in Figure~\ref{fig:more-vis-cityscapes} and Figure~\ref{fig:more-vis-ade20k}.

\end{document}